\pdfoutput=1
\documentclass{article} 
\usepackage[preprint]{my_style}
\usepackage{times}  
\usepackage{helvet} 
\usepackage{courier}  
\usepackage[hyphens]{url}  
\usepackage{graphicx} 
\usepackage{wrapfig}
\usepackage{amsfonts}       
\usepackage{amsmath}
\usepackage{mathrsfs}
\usepackage{amssymb}
\usepackage[boxed,ruled,commentsnumbered]{algorithm2e}
\usepackage{subcaption}
\usepackage{verbatim}
\newtheorem{assumption}{Assumption}
\newtheorem{lemma}{Lemma}
\newtheorem{theorem}{Theorem}

\newlength\savedwidth
\newcommand\whline{\noalign{\global\savedwidth\arrayrulewidth
		\global\arrayrulewidth 1.5pt}%
	\hline
	\noalign{\global\arrayrulewidth\savedwidth}}

\title{Deep Model-Based Reinforcement Learning via
	Estimated Uncertainty and Conservative Policy Optimization}

\author{Qi Zhou, Houqiang Li, Jie Wang\thanks{Corresponding author.}\\
	University of Science and Technology of China\\ 
	zhouqida@mail.ustc.edu.cn\\
	\{lihq, jiewangx\}@ustc.edu.cn
}
\begin{document}
	
	\maketitle

	\begin{abstract}
		Model-based reinforcement learning algorithms tend to achieve higher sample efficiency than model-free methods. However, due to the inevitable errors of learned models, model-based methods struggle to achieve the same asymptotic performance as model-free methods.
		In this paper, We propose a \textbf{P}olicy \textbf{O}ptimization method with \textbf{M}odel-\textbf{B}ased \textbf{U}ncertainty (POMBU)---a novel model-based approach---that can effectively improve the asymptotic performance using the uncertainty in Q-values. We derive an upper bound of the uncertainty, based on which we can approximate the uncertainty accurately and efficiently for model-based methods. We further propose an uncertainty-aware policy optimization algorithm that optimizes the policy conservatively to encourage performance improvement with high probability. This can significantly alleviate the overfitting of policy to inaccurate models. 
		Experiments show POMBU can outperform existing state-of-the-art policy optimization algorithms in terms of sample efficiency and asymptotic performance. Moreover, the experiments demonstrate the excellent robustness of POMBU compared to previous model-based approaches.
	\end{abstract}

	\section{Introduction}
	Model-free reinforcement learning has achieved remarkable success in sequential decision tasks, such as playing Atari games \cite{mnih2015human,hessel2018rainbow} and controlling robots in simulation environments \cite{lillicrap2015continuous,haarnoja2018soft}. However, model-free approaches require large amounts of samples, especially when using powerful function approximators, like neural networks. Therefore, the high sample complexity hinders the application of model-free methods in real-world tasks, not to mention data gathering is often costly. In contrast, model-based reinforcement learning is more sample efficient, as it can learn from the interactions with models and then find a near-optimal policy via models \cite{kocijan2004gaussian,deisenroth2011pilco,levine2014learning,nagabandi2018neural}. However, these methods suffer from errors of learned models, which hurt the asymptotic performance \cite{NIPS1996_1317,abbeel2006using}. Thus, compared to model-free methods, model-based algorithms can learn more quickly but tend to learn suboptimal policies after plenty of trials.
	
	Early model-based methods achieve impressing results using simple models, like linear models \cite{bagnell2001autonomous,levine2016end} and Gaussian processes \cite{kuss2004gaussian,deisenroth2011pilco}. However, these methods have difficulties in high-dimensional and non-linear environments due to the limited expressiveness of models. Recent methods use neural network models for better performance, especially for complicate tasks \cite{punjani2015deep,nagabandi2018neural}. Some methods further characterize the uncertainty in models via neural network ensembles \cite{rajeswaran2016epopt,kurutach2018model}, or Bayesian neural networks \cite{depeweg2016learning}. 
	Although the uncertainty in models improves the performance of model-based methods, recent research shows that these methods still struggle to achieve the comparable asymptotic performance to state-of-the-art model-free methods robustly \cite{1907.02057}. 
	
	Inspired by previous work that improves model-free algorithms via uncertainty-aware exploration \cite{o2017uncertainty}, we propose a theoretically-motivated algorithm to estimate the uncertainty in Q-values and apply it to the exploration of model-based reinforcement learning. Moreover, we propose to optimize the policy conservatively by encouraging a large probability of performance improvement, which is also informed by the estimated uncertainty. Thus, we use the uncertainty in Q-values to enhance both exploration and policy optimization in our model-based algorithm.
	
	Our contributions consist of three parts.
	
	First, we derive an upper bound of the uncertainty in Q-values and present an algorithm to estimate it. Our bound is tighter than previous work \cite{o2017uncertainty}, and our algorithm is feasible for deep model-based reinforcement learning, while many previous methods only focus on model-free cases \cite{osband2016deep,osband2017deep}, or assume simple models \cite{dearden1999model}.  
	
	Second, we propose to optimize the policy conservatively based on an estimated probability of performance improvement, which is estimated via the uncertainty in Q-values. We found the conservative policy optimization is useful to prevent the overfitting to the biased models. 
	
	Third, we propose a \textbf{P}olicy \textbf{O}ptimization method with \textbf{M}odel-\textbf{B}ased \textbf{U}ncertainty (POMBU), which combines our uncertainty estimation algorithm with the conservative policy optimization algorithm. Experiments show that POMBU achieves excellent robustness and can outperform state-of-the-art policy optimization algorithms. 
	
	\section{Background}
	A finite-horizon Markov decision process (MDP) $\mathcal{M}$ is defined by the tuple $(\mathcal{S}, \mathcal{A}, P, R, \rho, H)$. Here, $\mathcal{S}$ is a finite set of states, $\mathcal{A}$ is a finite set of actions, $P$ is a third-order tensor that denotes the transition probabilities, $R$ is a matrix that denotes the rewards, $\rho$ denotes the distribution of initial states and $H$ is the horizon length. More specifically, at the state $s$ and selecting the action $a$, $P_{sas^\prime} \in [0,1]$ is the probability of transitioning to the state $s^\prime$, and $R_{sa} \in [-R_{\mbox{max}}, R_{\mbox{max}}]$ is the obtained reward. We represent a posterior of MDPs as $(\varOmega,\mathcal{F},Pr)$, where $\varOmega$ is the sample space containing all possible MDPs, $\mathcal{F}$ is a $\sigma$-field consisting of subsets of $\varOmega$, and $Pr:\mathcal{F} \rightarrow [0,1]$ measures the posterior probability of MDPs. We assume that each MDP in $\varOmega$ is different from others only in terms of $P$ and $R$. In this case, $P$ is a random tensor and $R$ is a random matrix. For any random variable, matrix or tensor $X$, $\mathbb{E}_\mathcal{M}\left[X\right]$ and $\mathbb{D}_\mathcal{M}\left[X\right]$ denotes its expectation and variance respectively. When without ambiguity, we write $\mathbb{E}_\mathcal{M}\left[X\right]$ as $\bar{X}$ for short. For example,  $\bar{P}$ denotes $\mathbb{E}_\mathcal{M}\left[P\right]$ and $\bar{P}_{sas^\prime}$ denotes $\mathbb{E}_\mathcal{M}\left[P_{sas^\prime}\right]$.
	
	Let $\pi$ denotes a policy. $\pi_{sa}$ denotes the probability of taking the action $a$ at the state $s$. Considering the posterior of MDPs, the expected return $J_\pi$ is a random variable, which is defined by
	\begin{align*}
	&J_\pi = \mathbb{E}_{\tau\sim (\mathcal{M}, \pi)}\left[\sum_{h=1}^{H} R_{s^ha^h}\right].
	\end{align*}
	Here $\tau=(s^1,a^1,\dots,s^H,a^H)$ is a trajectory. $\tau\sim (\mathcal{M}, \pi)$ means that the trajectory is sampled from the MDP $\mathcal{M}$ under policy $\pi$. That is, $s^1$ is sampled from the initial state distribution $\rho$ of $\mathcal{M}$, $a^h$ is sampled with the probability $\pi_{s^ha^h}$ and $s^{h+1}$ is sampled with the probability $P_{s^ha^hs^{h+1}}$ in $\mathcal{M}$. Our goal is to find a policy maximizing $J^\pi$ in real environment. 
	
	Given an MDP $\mathcal{M}$, we define the corresponding state-action value function $Q$, the state value function $V$ and the advantage function $A$ as follow:
	\begin{align*}
	&V^h_\pi(s) = \mathbb{E}_{\tau\sim (\mathcal{M}, \pi)}\left[\sum_{l=h}^{H} R_{s^la^l}\left|s^h=s\right.\right],\\
	&Q^h_\pi(s,a) = \mathbb{E}_{\tau\sim (\mathcal{M}, \pi)}\left[\sum_{l=h}^{H} R_{s^la^l}\left|s^h=s, a^h=a\right.\right],\\
	&A^h_\pi(s,a) = Q^h_\pi(s,a) -  V^h_\pi(s).
	\end{align*} 
	
	When the policy $\pi$ is fixed, we write $V^h_\pi(s)$, $Q^h_\pi(s,a)$ and $A^h_\pi(s,a)$ as ${V}_{s}^h$, ${Q}_{sa}^h$ and ${A}^h_{sa}$ respectively for short. In this case, for any time-step $h$, ${V}_{s}^h$, ${Q}_{sa}^h$ and ${A}^h_{sa}$ are random variables mapping $\varOmega$ to $\mathbb{R}$. Hence, $V^h$ is a random vector. $Q^h$ and $A^h$ are random matrices.

	\section{Uncertainty Estimation}
	In this section, we consider a fixed policy $\pi$. Similarly to the uncertainty Bellman equation (UBE) \cite{o2017uncertainty}, we regard the standard deviations of Q-values as the uncertainty. In this section, we derive an upper bound of $\mathbb{D}_\mathcal{M}\left[Q^h_{sa}\right]$ for each $s, a, h$, and prove that our upper bound is tighter than that of UBE.
	Moreover, we propose an uncertainty estimation algorithm for deep model-based reinforcement learning and discuss its advantages. We provide related proofs in Appendix A.1-A.4.
	
	\subsection{Upper Bound of Uncertainty in Q-values}
	To analyze the uncertainty, we first make two assumptions. 
	
	\begin{assumption}\label{assumption1}
		Each MDP in $\varOmega$ is a directed acyclic graph.
	\end{assumption}
	This assumption is common \cite{osband2014generalization,o2017uncertainty}. It means that the agent cannot visit a state more than twice within the same episode. This assumption is weak because each finite horizon MDP violating the assumption can be converted into a similar MDP that satisfying the assumption \cite{o2017uncertainty}.
	
	\begin{assumption}\label{assumption2}
		The random vector $R_{s_1}$ and the random matrix $P_{s_1}$ are independent of $R_{s_2}$ and $P_{s_2}$ if $s_1 \neq s_2$.
	\end{assumption}
	This assumption is used in the derivation of UBE \cite{o2017uncertainty}. It is consistent with the trajectory sampling strategies used in recent model-based algorithms \cite{chua2018deep,kurutach2018model}, which sample a model from the ensemble of models independently per time step to predict the next state and reward.
	
	First, we derive an inequation from these assumptions. 
	
	\begin{lemma}\label{lemma1}
		\label{LEMMA1} Under Assumption 1 and 2, for any $s \in \mathcal{S}$ and $a \in \mathcal{A}$, we have
		\begin{align*}
		&\mathbb{D}_\mathcal{M}\left[Q^h_{sa}\right] \leq u^{h}_{sa} + \sum_{s^\prime, a^\prime}\pi_{s^\prime a^\prime}\bar{P}_{sas^\prime}\mathbb{D}_\mathcal{M}\left[Q^{h+1}_{s^\prime a^\prime}\right],\\
		&\mbox{where } u^h_{sa} = \mathbb{D}_{\mathcal{M}}\left[R_{sa} + \sum_{s^\prime}P_{sas^\prime}\bar{V}^{h+1}_{s^\prime}\right].
		\end{align*}
	\end{lemma}
	We consider $u^h_{sa}$ as a local uncertainty, because we can compute it locally with $\bar{V}$.
	
	Then, we can derive our main theorem from this lemma. 
	
	\begin{theorem}\label{theorem1}
		Under Assumption 1 and 2, for any policy $\pi$, there exists a unique solution $U$ satisfying the following equation:
		\begin{align}\label{ours_UBE}
		&U^h_{sa} = u^h_{sa} + \sum_{s^\prime, a^\prime}\pi_{s^\prime a^\prime}\bar{P}_{sas^\prime}U^{h+1}_{s^\prime a^\prime}
		\end{align}
		for any $(s,a)$ and $h=1,2,\dots,H$, where $U^{H+1}=\textbf{0}$, and furthermore $U \geq \mathbb{D}_\mathcal{M}\left[Q\right]$ pointwise.
	\end{theorem}
	Theorem 1 means that we can compute an upper bound of $\mathbb{D}_\mathcal{M}\left[Q^h_{sa}\right]$ by solving the Bellman-style equation (\ref{ours_UBE}). 
	
	Moreover, we provide the following theorem to show the convergence when computing $U$ iteratively.
	\begin{theorem}\label{theorem2}
		For arbitrary $\left(U\right)_1$, if
		$$
		\left(U^h_{sa}\right)_{i+1} = \left(u^h_{sa}\right)_i + \sum_{s^\prime, a^\prime}\pi_{s^\prime a^\prime}\bar{P}_{sas^\prime}\left(U^{h+1}_{s^\prime a^\prime}\right)_i,
		$$
		for any $(s,a)$, $h=1,2,\dots,H$ and $i \geq 1$, where $\left(U^{H+1}\right)_i=\textbf{0}$ and $\left(u\right)_i$ converges to $u$ pointwise, we have $\left(U\right)_i$ converges to $U$ pointwise.
	\end{theorem}
	Theorem 2 shows that we can solve the equation (\ref{ours_UBE}) iteratively if the estimated local uncertainty is inaccurate per update but converges to the correct value, which is significant when we use an estimated $\bar{V}$ to compute the uncertainty.
	
	As $U^h_{sa}$ is an upper bound of $\mathbb{D}_\mathcal{M}\left[Q^h_{sa}\right]$, $\sqrt{U^h_{sa}}$ is an upper bound of the uncertainty in $Q^h_{sa}$. We use the upper bound to approximate the uncertainty in our algorithm similarly to UBE. We need to analyze the accuracy of our estimates. 
	
	Here, we compare our upper bound $U$ with that of UBE under the same assumptions, and hence we need to make an extra assumption used in UBE. 
	\begin{assumption}
		$R_{s}$ is independent of $P_{s}$ for any $s \in \mathcal{S}$.
		\label{mu_ass}
	\end{assumption}
	This assumption is not used to derive our upper bound of the uncertainty but is used in UBE. Under the assumption 2 and 3, we have $R$ is independent of $P$. 
	
	The upper bound $B$ derived in UBE satisfies 
	$$
	B^h_{sa} = \nu^h_{sa} + \sum_{s^\prime, a^\prime}\pi_{s^\prime a^\prime}\bar{P}_{sas^\prime}B^{h+1}_{s^\prime a^\prime},
	$$
	where $\nu^h_{sa}= \left(Q_{\mbox{max}}\right)^2 \sum_{s^\prime}\mathbb{D}_{\mathcal{M}}\left[P_{sas^\prime}\right]/\bar{P}_{sas^\prime} +  \mathbb{D}_{\mathcal{M}}\left[R_{sa}\right]$.
	Here, $Q_{\mbox{max}}$ is an upper bound of all $|Q^h_{sa}|$ for any $s,a,h$ and MDP. For example, we can regard $HR_{\mbox{max}}$ as $Q_{\mbox{max}}$.
	
	\begin{theorem}\label{tighter}
		Under the assumption 1, 2 and 3, $U^h_{sa}$ is a tighter upper bound of $\mathbb{D}_\mathcal{M}\left[Q^h_{sa}\right]$ than $B^h_{sa}$.
	\end{theorem}
	This theorem means that our upper bound $U$ is a more accurate estimate of the uncertainty in Q-values than the upper bound $B$ derived in UBE.

	\subsection{Uncertainty Estimation Algorithm}\label{ue}
	First, we characterizes the posterior of MDPs approximatively using a deterministic model ensemble (please refer to the Section \ref{se5} for the details of training models). A deterministic ensemble is denoted by $(f_{w_1}, \dots, f_{w_K})$. Here, for any $i=1,\dots,K$, $f_{w_i}: \mathcal{S}\times\mathcal{A} \rightarrow \mathcal{S}\times\mathbb{R}$ is a single model that predicts the next state and the reward, and $w_i$ is its parameters. We define a posterior probability of MDPs by
	$$
	Pr\{P_{sas^\prime} = 1, R_{sa}=x\} = \frac{1}{K}\sum_{i=1}^{K} \mbox{\bfseries eq}\left((s^\prime, x), f_{w_i}(s,a)\right),
	$$
	where \textbf{eq} is defined by
	$$\mbox{\bfseries eq}\left((s_1, x_1), (s_2, x_2)\right) = \\
	\begin{cases}
	1, &\mbox{if } s_1 = s_2 \mbox{ and } x_1 = x_2,\\
	0, &\mbox{otherwise}.
	\end{cases}
	$$ 
	
	Then, we can construct an MDP $\hat{\mathcal{M}}$ defined according to the posterior of MDPs, such that its transition tensor $\hat{P}$ is equal to $\bar{P}$ and its reward matrix $\hat{R}$ is equal to $\bar{R}$. Hence, the state value matrix $\hat{V}$ of the MDP $\hat{\mathcal{M}}$ is equal to $\bar{V}$. 
	
	Moreover, we use a neural network $\tilde{V}_\phi: \mathcal{S} \rightarrow \mathbb{R}$ to predict $\hat{V}^h_s$ for any state $s$ and time step $h$, which is equivalent to predicting $\bar{V}$. We train $\tilde{V}_\phi$ by minimizing $\ell_2$ loss function  
	\begin{align}
	L_v(\phi) =\mathbb{E}_{\tau\sim(\hat{\mathcal{M}}, \pi)}\left[\frac{1}{H}\sum_{h=1}^{H}\left\|\tilde{V}_\phi(s^h)- \sum_{l=h}^{H}\hat{R}_{s^la^l}\right\|_2^2\right].\label{value}
	\end{align}
	
	Finally, given an imagined trajectory $\tau$ sampled from $\hat{\mathcal{M}}$ under $\pi$, we can estimate the uncertainty in Q-values via the algorithm \ref{alg1}. Note that for long-horizon tasks, we can introduce a discount factor $\gamma$ similarly to previous work \cite{o2017uncertainty}. The modified uncertainty estimation method can be found in Appendix B.

	\begin{algorithm}[ht]
		\caption{Uncertainty Estimation for Q-values}
		\label{alg1}
		\SetAlgoNoLine 
		\SetKwInOut{Input}{\textbf{Input}}\SetKwInOut{Output}{\textbf{Output}}
		\Input{A approximate value function $\tilde{V}_\phi$; An ensemble model $\{f_{w_1}, f_{w_2}, \dots, f_{w_K} \}$; A trajectory $\tau=(s_1, a_1, \dots, s_H, a_H)$\;\\}
		\Output{Estimates of $(\sqrt{U^1_{s_1a_1}}, \dots, \sqrt{U^H_{s_Ha_H}})$;\\}
		\BlankLine
		$D^{H+1}=0$\;
		\For{$i = H,H-1,\dots,1$}{
			\For{$j = 1,2,\dots,K$}{
				$(s_j, r_j) = f_{w_j}(s^i,a^i)$\;
				$q_j = r_j + \tilde{V}_\phi(s_j)$\;
			}
			$q = \frac{1}{K}\sum_{j=1}^{K}q_j$\;
			$d^i = \frac{1}{K}\sum_{j=1}^{K}(q_j-q)^2$\;
			$D^{i} = d^i + D^{i+1}$\;
		}
		\Return $(\sqrt{D^1}, \sqrt{D^2}, \dots, \sqrt{D^H})$;
	\end{algorithm}
	
	\subsection{Discussion}\label{benefits}
	In this part, we discuss some advantages of our algorithm to estimate the uncertainty in Q-values.
	\subsubsection{Accuracy}
	Based on the Theorem \ref{tighter}, our upper bound of the uncertainty is tighter than that of UBE, which means a more accurate estimation. Intuitively, our local uncertainty depends on $\bar{V}^h$ while that of UBE depends on $Q_{\mbox{max}}$. Therefore, our local uncertainty has a weaker dependence on $H$ and can provide a relatively accurate estimation for long-horizon tasks (see an example in Appendix C). Moreover, considering an infinite set of states, our method ensures the boundedness of the local uncertainty because $\bar{V}^h$ and $R$ are bounded. Therefore, our method has the potential to apply to tasks with continuous action spaces.
	
	\subsubsection{Applicability for Model-Based Methods}
	Our method to estimate the uncertainty in Q-values is effective for model-based reinforcement learning. In model-based cases, estimated Q-values are highly dependent on the models. Our method considers the model when computing the local uncertainty, while most of the existing methods estimate the uncertainty directly via the real-world samples regardless of the models. Ignoring models may lead to bad estimates of uncertainty in model-based cases. For example, the uncertainty estimated by a count-based method \cite{bellemare2016unifying,ostrovski2017count} tends to decrease with the increase of the number of samples, while the true uncertainty keeps high even with a large amount of samples when modeling a complicate MDP using a simple model.
	
	\subsubsection{Computational Cost}
	Our method is much more computationally cheap compared with estimating the uncertainty via the empirical standard deviation of $Q_{sa}^h$. When MDP is given, estimating $Q_{sa}^h$ requires plenty of virtual samples. Estimating the empirical standard deviation requires estimating $Q_{sa}^h$ for several MDPs. Previous work reduces the computational cost by learning an ensemble of Q functions \cite{buckman2018sample}. However, training an ensemble of Q functions requires higher computational overhead than training a single neural network $\tilde{V}_\phi$.
	
	\subsubsection{Compatibility with Neural Networks}
	Previous methods that estimate uncertainty for model-based methods always assume simple models, like Gaussian processes \cite{deisenroth2011pilco,dearden1999model}. Estimating uncertainty using Theorem \ref{theorem1} only requires that the models can represent a posterior. This makes our method compatible with neural network ensembles and Bayesian neural networks. For instance, we propose Algorithm \ref{alg1} with an ensemble of neural networks.
	
	\subsubsection{Propagation of Uncertainty}
	As discussed in previous work \cite{osband2018randomized}, Bellman equation implies the high dependency between Q-values. Ignoring this dependence will limit the accuracy of the estimates of uncertainty. Our method considers the dependency and propagates the uncertainty via a Bellman-style equation.
	
	\section{Conservative Policy Optimization}
	In this section, we first introduce surrogate objective and then modify it via uncertainty. The modified objective leads to conservative policy optimization because it penalizes the update in the high-uncertainty regions. $\pi_{\theta}: \mathcal{S}\times\mathcal{A}\rightarrow [0,1]$ denotes a parameterized policy, and $\theta$ is its parameters. $\pi_{\theta}(a|s)$ is the probability of taking action $a$ at state $s$. 
	\subsection{Surrogate Objective}
	Recent reinforcement learning algorithms, like Trust Region Policy Optimization (TRPO) \cite{schulman2015trust}, Proximal Policy Optimization (PPO) \cite{schulman2017proximal}, optimize the policy based on surrogate objective. We rewrite the surrogate objective in TRPO and PPO as follow: 
	\begin{align*}
	L_{sr}(\theta) =  \mathbb{E}_{\tau\sim (\mathcal{M}, \pi_{\theta_{\text{old}}})}\left[\sum_{h=1}^{H}r_\theta(s^h,a^h)A^h_{{\text{old}}}(s^h,a^h)\right],
	\end{align*}
	where $\theta_{\text{old}}$ are the old policy parameters before the update, $A_{{\text{old}}}$ is the advantage function of $\pi_{\theta_{\text{old}}}$ and 
	$$
	r_\theta(s,a) = \frac{\pi_{\theta}(a|s)}{\pi_{\theta_{\text{old}}}(a|s)}.
	$$
	Previous work has proven the surrogate objective is the first order approximation to $J(\pi_{\theta}) - J(\pi_{\theta_{\text{old}}})$ when ${\theta}$ is around ${\theta_{\text{old}}}$ \cite{schulman2015trust,kakade2002approximately}. That is, for any $\theta_{\text{old}}$, we have the following theorem:
	
	\begin{theorem}\label{appro}
		\begin{align*}
		\left.L_{sr}(\theta)\right|_{\theta=\theta_{\text{old}}} &= \left.J(\pi_{\theta}) - J(\pi_{\theta_{\text{old}}})\right|_{\theta=\theta_{\text{old}}},\\
		\left.\nabla_\theta L_{sr}(\theta)\right|_{\theta=\theta_{\text{old}}} &= \nabla_\theta\left.\left(J(\pi_{\theta}) - J(\pi_{\theta_{\text{old}}})\right)\right|_{\theta=\theta_{\text{old}}}
		\end{align*}
	\end{theorem}
	(see proof in Appendix A.5). Therefore, maximizing $L_{sr}(\theta)$ can maximize $J(\pi_{\theta})$ approximately when ${\theta}$ is around ${\theta_{\text{old}}}$. 
	
	\subsection{Uncertainty-Aware Surrogate Objective}\label{ob}

	To prevent the overfitting of the policy to inaccurate models, we introduce the estimated uncertainty in Q-values into the surrogate objective. 
	
	First, we need to estimate $Pr\{J(\pi_{\theta})>J(\pi_{\theta_{\text{old}}})\}$, which means the probability that the new policy outperforms the old one. Because of Theorem \ref{appro}, $Pr\{L_{sr}(\theta)>0\}$ can approximate $Pr\{J(\pi_{\theta})>J(\pi_{\theta_{\text{old}}})\}$.
	We assume that a Gaussian can approximate the distribution of $L_{sr}(\theta)>0$. Thus, $F\left(\mathbb{E}_\mathcal{M}\left[L_{sr}(\theta)\right]/\sqrt{\mathbb{D}_\mathcal{M}\left[L_{sr}(\theta)\right]}\right)$ is approximately equal to $Pr\{L_{sr}(\theta)>0\}$, where $F$ is the probability distribution function of standard normal distribution.
	
	Then, we need to construct an objective function for optimization. Here, we aims to find a new $\theta$ with a large $F\left(\mathbb{E}_\mathcal{M}\left[L_{sr}(\theta)\right]/\sqrt{\mathbb{D}_\mathcal{M}\left[L_{sr}(\theta)\right]}\right)$. As $F$ is monotonically increasing, we can maximize $\mathbb{E}_\mathcal{M}\left[L_{sr}(\theta)\right]$ while minimize $\sqrt{\mathbb{D}_\mathcal{M}\left[L_{sr}(\theta)\right]}$. Therefore, we can maximize
	\begin{align}
	\mathbb{E}_\mathcal{M}\left[L_{sr}(\theta)\right] -\alpha\sqrt{\mathbb{D}_\mathcal{M}\left[L_{sr}(\theta)\right]},\label{overall}
	\end{align}
	where $\alpha \geq 0$ is a hyperparameter. 
	
	Moreover, we need to estimate the expectation and the variance of the surrogate objective. Because $L_{sr}(\theta)$ is equal to
	$$
	{\mathbb{E}_{\tau\sim({\mathcal{M}},\pi_{\theta_{\text{old}}})}\left[\sum_{h=1}^{H}\left(r_\theta(s^h,a^h)-1\right){Q^h_{{\text{old}}}}(s^h,a^h) \right]},
	$$
	we can approximate $\mathbb{E}_\mathcal{M}\left[L_{sr}(\theta)\right]$ and $\sqrt{\mathbb{D}_\mathcal{M}\left[L_{sr}(\theta)\right]}$ as $L_{\mbox{exp}}(\theta)$ and $L_{\mbox{std}}(\theta)$ respectively, where
	\begin{align}
	&L_{\mbox{exp}}(\theta) =   {\mathbb{E}_{\tau\sim(\hat{\mathcal{M}},\pi_{\theta_{\text{old}}})}\left[\sum_{h=1}^{H}r_\theta(s^h,a^h)\bar{A}^h_{{\text{old}}}(s^h,a^h)\right]},\\
	&L_{\mbox{std}}(\theta) = {\mathbb{E}_{\tau\sim(\hat{\mathcal{M}},\pi_{\theta_{\text{old}}})}\left[\sum_{h=1}^{H}\left|r_\theta(s^h,a^h)-1\right|\sqrt{D^h}\right]}.\label{var}
	\end{align}
	Here $\hat{\mathcal{M}}$ is defined in Section \ref{ue} using a learned ensemble, $\bar{A}^h_{{\text{old}}}(s^h,a^h)$ can be approximated by $\sum_{l=h}^{H}\hat{R}_{s^la^l} -\tilde{V}_\phi(s^h)$, and $D^h$ is computed by Algorithm \ref{alg1}.
	
	However, policy optimization without trust region may lead to unacceptable bad performance \cite{schulman2015trust}. Thus, we clip $L_{\mbox{exp}}(\theta)$ similarly to PPO. That is,
	\begin{align}
	L_{\mbox{clip}}(\theta) = {\mathbb{E}_{\tau\sim(\hat{\mathcal{M}},\pi_{\theta_{\text{old}}})}\left[\sum_{h=1}^{H}\hat{r}_\theta(s^h,a^h)\bar{A}^h_{{\text{old}}}(s^h,a^h)\right]}.\label{ex}
	\end{align}
	Here, we define $\hat{r}_\theta(s^h,a^h)$ as
	$$
	\begin{cases}
	\max(1-\epsilon, {r}_\theta(s^h,a^h)), &\mbox{if }\bar{A}^h_{{\text{old}}}(s^h,a^h) \leq 0,\\
	\min(1+\epsilon, {r}_\theta(s^h,a^h)), &\mbox{if }\bar{A}^h_{{\text{old}}}(s^h,a^h) > 0,\\
	\end{cases}
	$$
	in which $\epsilon>0$ is a hyperparameter.
	
	Finally, we obtain the modified surrogate objective
	$$L_\pi(\theta) = L_{\mbox{clip}}(\theta) - \alpha L_{\mbox{std}}(\theta).$$ 
	Note that, the main difference of our objective from PPO is the uncertainty penalty $L_{\mbox{std}}(\theta)$. This penalty limits the ratio changes $|r_\theta(s^h,a^h)-1|$ in high-uncertainty regions. Therefore, this objective is uncertainty-aware and leads to a conservative update.

	\begin{algorithm}[ht]
		\caption{POMBU}
		\label{alg2}
		\SetAlgoNoLine 
		\BlankLine
		Initialize an ensemble $\{\tilde{f}_{w_1}\dots \tilde{f}_{w_K}\}$ and a policy $\pi_{\theta}$\;
		Initialize a value function $\tilde{V}_\phi$\;
		Initialize the dataset $S$ as a empty set\;
		Sample $N$ trajectories using $\pi_{\theta}$\;
		Add the sampled transitions to $S$\;
		\Repeat{\rm $\pi_{\theta}$ performs well in the real environment
		}{
			Train the ensemble $\{\tilde{f}_{w_1}\dots \tilde{f}_{w_K}\}$ using $S$\;
			\For{$i = 1,2,\dots,M$}{
				Sample virtual trajectories from $\hat{\mathcal{M}}$ using $\pi_{\theta}$\;
				Train $\tilde{V}_\phi$ by minimizeing $L_v(\phi)$\;
				Train $\pi_{\theta}$ by maximizing  $L_\pi(\theta)$\;
			}
			\For{$i = 1,2,\dots,N$}{
				Sample virtual trajectories from $\hat{\mathcal{M}}$ using $\pi_{\theta}$\;
				Train an exploration policy $\pi_{\text{exp}}$ based on $\pi_{\theta}$\;
				Collect real-world trajectories using $\pi_{\text{exp}}$\;
				Add the sampled transitions to $S$\;
			}
		}
	\end{algorithm}

	\section{Algorithm}\label{se5}

	In this section, we propose a \textbf{P}olicy \textbf{O}ptimization method with \textbf{M}odel-\textbf{B}ased \textbf{U}ncertainty (POMBU) in Algorithm \ref{alg2}. We detail each stage of our algorithm as following.
	\subsubsection{Exploration Policy}
	We train a set of exploration policies by maximizing the $L_{\mbox{clip}}(\theta)$. Different policies are trained with different virtual trajectories. To explore the unknown, we replace $\bar{A}^h_{{\text{old}}}(s^h,a^h)$ with $\bar{A}^h_{{\text{old}}}(s^h,a^h) + \beta\sqrt{D^h}$ in the equation (\ref{ex}). Here, $\beta \geq 0$ controlling the exploration to high-uncertainty regions. 
	\subsubsection{Model Ensemble}
	To predict the next state, a single neural network in the ensemble outputs the change in state and then adds the change to the current state \cite{kurutach2018model,nagabandi2018neural}. To predict the reward, we assume the reward in real environment is computed by a function $\mu$ such that $R_{s^ha^h} = \mu(s^h,a^h,s^{h+1})$, which is commonly true in many simulation control tasks. Then, we can predict the reward via the predicted next state. We train the model by minimizing $\ell_2$ loss similarly to previous work \cite{kurutach2018model,nagabandi2018neural} and optimize the parameter using Adam \cite{kingma2014adam}. Different models are trained with different train-validation split.
	\subsubsection{Policy Optimization}
	We use a Gaussian policy whose mean is computed by a forward neural network and standard deviation is represented by a vector of parameters. We optimizing all parameters by maximizing $L_\pi(\theta)$ via Adam.

	\section{Experiments}
	In this section, we fist evaluate our uncertainty estimation method. Second, we compare POMBU to state-of-the-arts. Then, we show how does the estimated uncertainty work by ablation study. Finally, we analyze the robustness of our method empirically. In the following experiments, we report the performance averaged over at least three random seeds. Please refer to Appendix D for the details of experiments. The source code and appendix of this work is available at \url{https://github.com/MIRALab-USTC/RL-POMBU}. 
	
	\subsection{Effectiveness of Uncertainty Estimation}\label{uncertainty_exp}
	
	\begin{figure*}[ht]
		\centering
		\includegraphics[width=.90\textwidth]{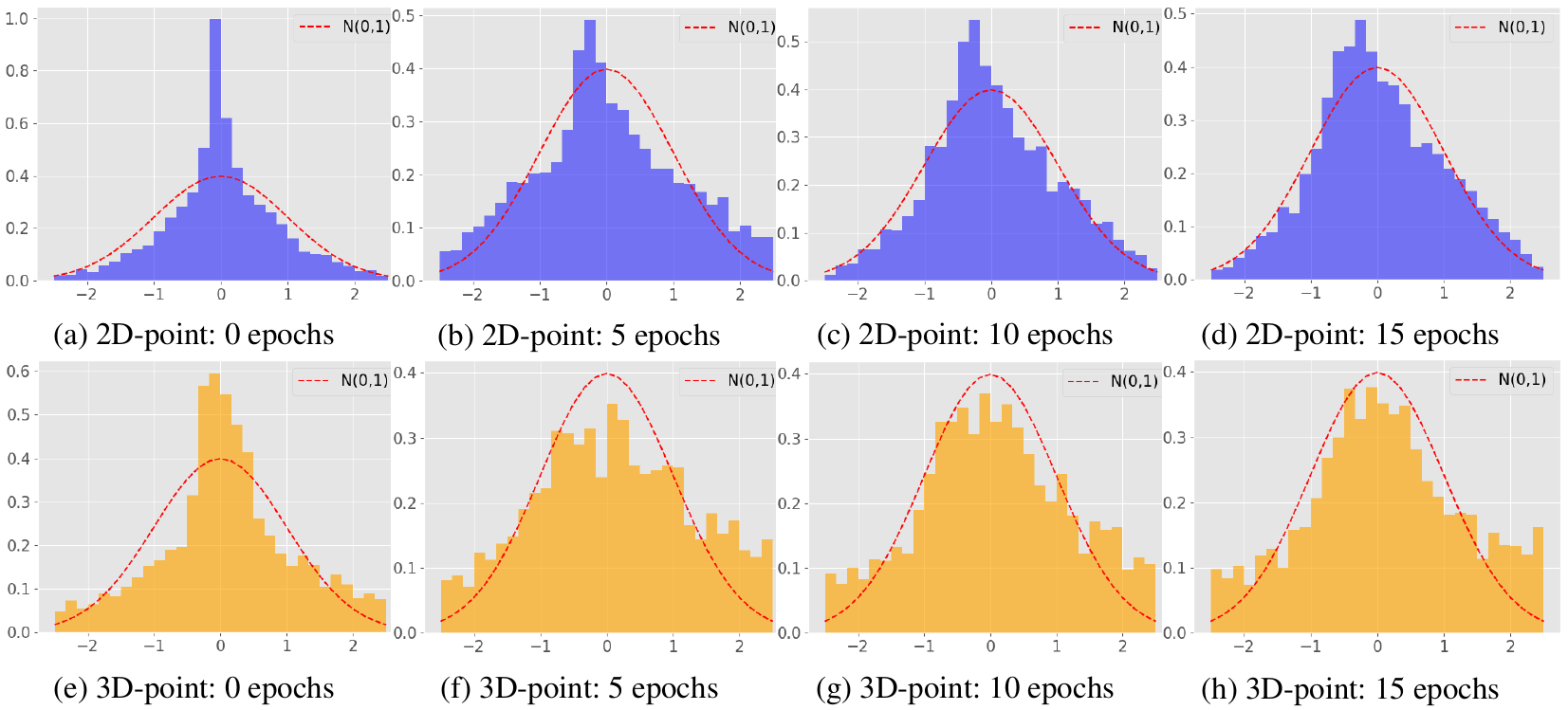}
		\caption{Frequency histograms of the ratios of errors to uncertainties after different numbers of epochs (training $\tilde{V}_\phi$). The red dotted line means the probability density function of the standard normal distribution.}
		\label{fig:ratio}
	\end{figure*}
	We evaluate the effectiveness of our uncertainty estimation method in two environments: 2D-point and 3D-point. These environments have continuous state spaces and continuous action spaces. First, we train an ensemble model of the environment and sample state-action pairs from the model using a deterministic policy. Then, we estimate the Q-values of these pairs via the means of virtual returns (computed using the models), and estimate the uncertainty using the algorithm \ref{alg1}. Finally, we compute the real Q-values using the return in real world, compute the ratios of errors to the estimated uncertainties, and count the frequencies of these ratios to draw Figure \ref{fig:ratio}. This figure shows the distribution of ratios is similar to a standard normal distribution after sufficient training of $\tilde{V}_\phi$, which demonstrates the accuracy of the estimated uncertainty.

	\subsection{Comparison to State-of-the-Arts}
	We compare POMBU with state-of-the-art policy optimization algorithms in four continuous control tasks of Mujoco \cite{todorov2012mujoco}: Swimmer, HalfCheetah, Ant, and Walker2d. Our method and our baselines optimize a stochastic policy to complete the tasks. Our baselines include: soft actor critic (SAC) \cite{haarnoja2018soft}; proximal policy optimization (PPO); stochastic lower bounds optimization (SLBO) \cite{luo2018algorithmic}; model-ensemble trust region policy optimization (METRPO) \cite{kurutach2018model}. To show the benefits of using uncertainty in model-based reinforcement learning, we also compare POMBU to model-ensemble proximal policy optimization (MEPPO), which is equivalent to POMBU when $\alpha=0$ and $\beta=0$. We evaluate POMBU with $\alpha=0.5$ and $\beta=10$ for all tasks. 
	
\begin{figure}[ht]
		\centering
		\includegraphics[width=.85\textwidth]{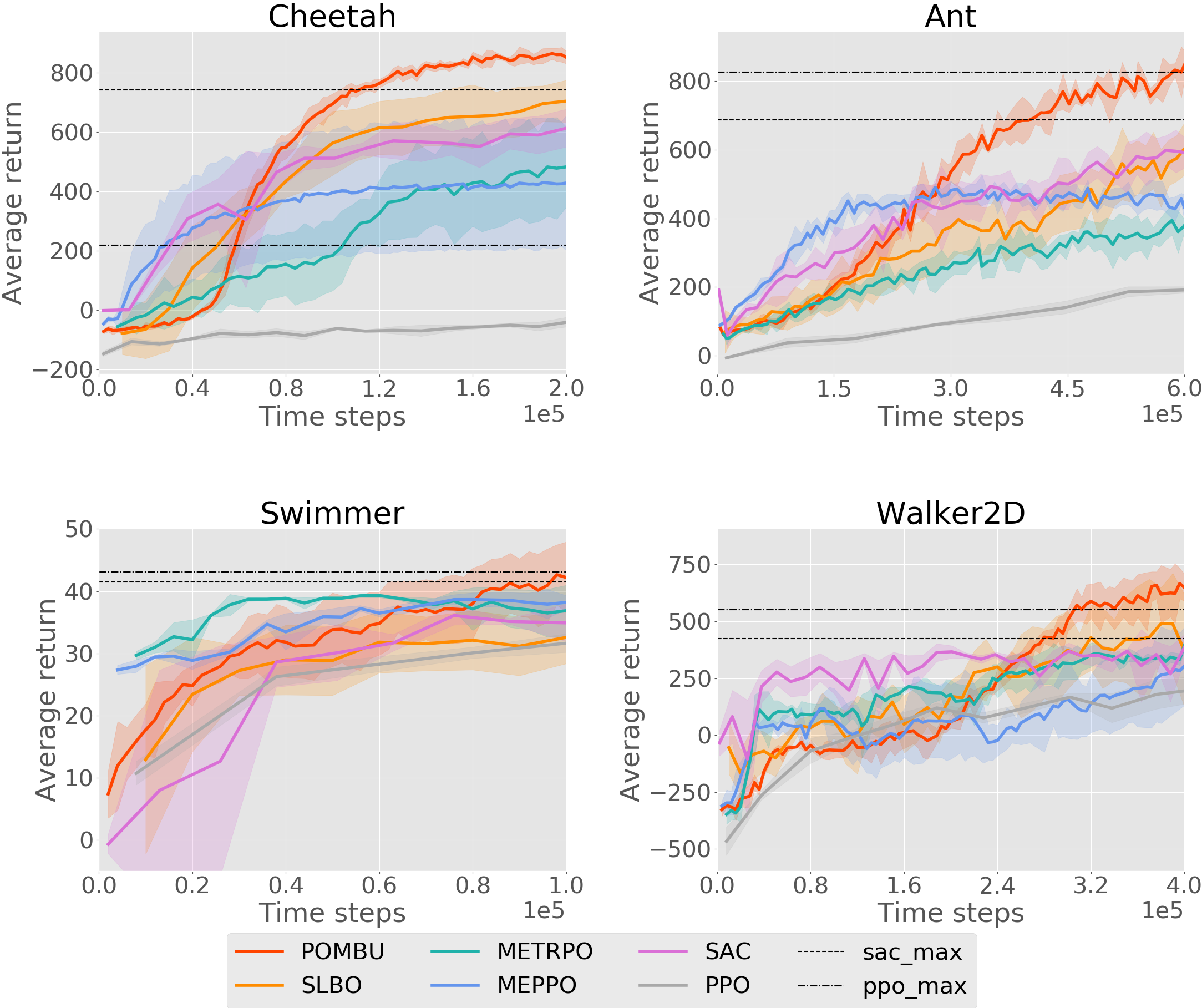}
		\caption{The training curve of our method and baselines. The horizons of all tasks are 200. The number of total steps is selected to ensure most model-based algorithms converge. We train the policy via PPO and SAC with at least 1 million samples and report the best averaged performance as "max". }
		\label{fig:compare}
	\end{figure}

	The result is shown in Figure \ref{fig:compare}. The solid curves correspond to the mean and the shaded region corresponds to the empirical standard deviation.
	It shows that POMBU achieves higher sample efficiency and better final performance than baselines, which highlights the great benefits of using uncertainty. Moreover, POMBU achieves comparable asymptotic performances with PPO and SAC in all tasks.
	
	We also provide Table \ref{sum} that summarizes the performance, estimated wall-clock time and the number of used imagined samples and real-world samples in the HalfCheetah task (H=200). Compared to MEPPO, the extra time used in POMBU is small (time: $10.17 \rightarrow 12.05$), while the improvement is significant (mean: $449 \rightarrow 852$; standard deviation: $226 \rightarrow 21$). Compared to SAC, POMBU achieve higher performance with about 5 times less real-world samples. Moreover, in our experiments, the total time to compute the uncertainty (not include the time to train $\tilde{V}_\phi$) is about 1.4 minutes, which is ignorable compared with the overall time.
	
	We further compare POMBU with state-of-the-art model-based algorithms in long-horizon tasks. The compared algorithms include model-based meta policy optimization (MBMPO) \cite{clavera2018model}, probabilistic ensemble with trajectory sampling (PETS) \cite{chua2018deep} and stochastic ensemble value expansion (STEVE) \cite{buckman2018sample} in addition. We directly use some of the results given by Tingwu Wang \cite{1907.02057}, and summarize all results in Table \ref{long}. The table shows that POMBU achieves comparable performance with STEVE and PETS, and outperforms other model-based algorithms. It demonstrates that POMBU is also effective in long-horizon tasks.
	
	\begin{table*}[ht]
		\centering
		\begin{tabular}{c  c  c  c  c  c  c  c  c}
			\whline
			& \textbf{POMBU} & \textbf{MEPPO} & \textbf{METRPO} & \textbf{SLBO} & \textbf{SAC} & \textbf{PPO} & \textbf{SAC\_max} & \textbf{PPO\_max}\\ \hline
			\textbf{Time (h)} & 12.05 & 10.17 & 6.35 & 3.91 &0.87 &0.04 &4.18 &0.19 \\ 
			\textbf{Imagined} &1.2e8 &8e7 &5e7 &1e7 &0 &0 &0 &0 \\ 
			\textbf{Real-world} &2e5 &2e5 &2e5 &2e5 &2e5 &2e5 &9.89e5 &9.78e5 \\ 
			\textbf{Return} &$852\pm21$ &$449\pm226$ &$483\pm136$ &$704\pm70$ &$615\pm64$ &$-38\pm16$ &$741\pm88$ &$218\pm63$  \\ 
			\whline
		\end{tabular}
		
		\caption{The performance, estimated wall-clock time and the number of used imagined samples and real-world samples in the HalfCheetah task (H=200). We conduct all experiments with  one GPU Nvidia GTX 2080Ti. }
		\label{sum}
		
	\end{table*}

	\begin{table*}[ht]
		\centering
		\begin{tabular}{c  c  c  c  c  c c}
			\whline
			\textbf{Environment } & \textbf{POMBU} & \textbf{STEVE} & \textbf{MBMPO} & \textbf{SLBO} & \textbf{METRPO} & \textbf{PETS} \\ \hline
			\textbf{Ant} & $\mathbf{2010\pm91}$ & $552\pm190$ & $706\pm147$ & $728\pm123$ & $282\pm18$ & $1852\pm141$\\ 
			\textbf{HalfCheetah} & $3672\pm8$ & $\mathbf{7965\pm1719}$ & $3639\pm1186$ & $1098\pm166$ & $2284\pm900$ & $2795\pm880$\\ 
			\textbf{Swimmer} & $\mathbf{144.4\pm22.6}$ & $\mathbf{149\pm81}$ & $85.0\pm98.9$ & $41.6\pm18.4$ & $30.1\pm9.7$ & $22.1\pm25.2$\\ 
			\textbf{Walker2d} & $-565\pm129$ & $-26\pm328$ & $-1546\pm217$ & $-1278\pm428$ & $-1609\pm658$ & $\mathbf{260\pm537}$\\ 
			\whline
		\end{tabular}
		
		\caption{The performance of 200k time-step training. The horizons of all environments are 1000. }
		\label{long}
		
	\end{table*}

	\subsection{Ablation Study}
	\begin{wrapfigure}{r}{0.54\textwidth}
		\centering
		\vspace{-35pt}
		\includegraphics[width=.48\columnwidth]{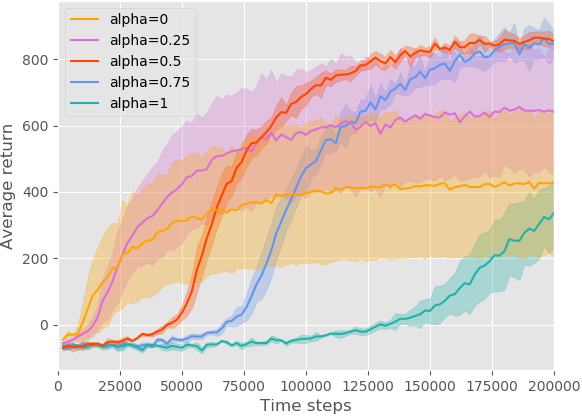}
		\caption{The development of the average return during training with different $\alpha$ in the Cheetah task (H=200).}
		\label{fig:abalation}
	\end{wrapfigure}
	We provide an ablation study to show how the uncertainty benefits the performance. In our algorithm, we employ the uncertainty in policy optimization (controlled by $\alpha$) and exploration (controlled by $\beta$). Therefore, we compare the performance with different $\alpha$ and $\beta$.

\begin{figure}[ht]
		\centering
		\includegraphics[width=.85\columnwidth]{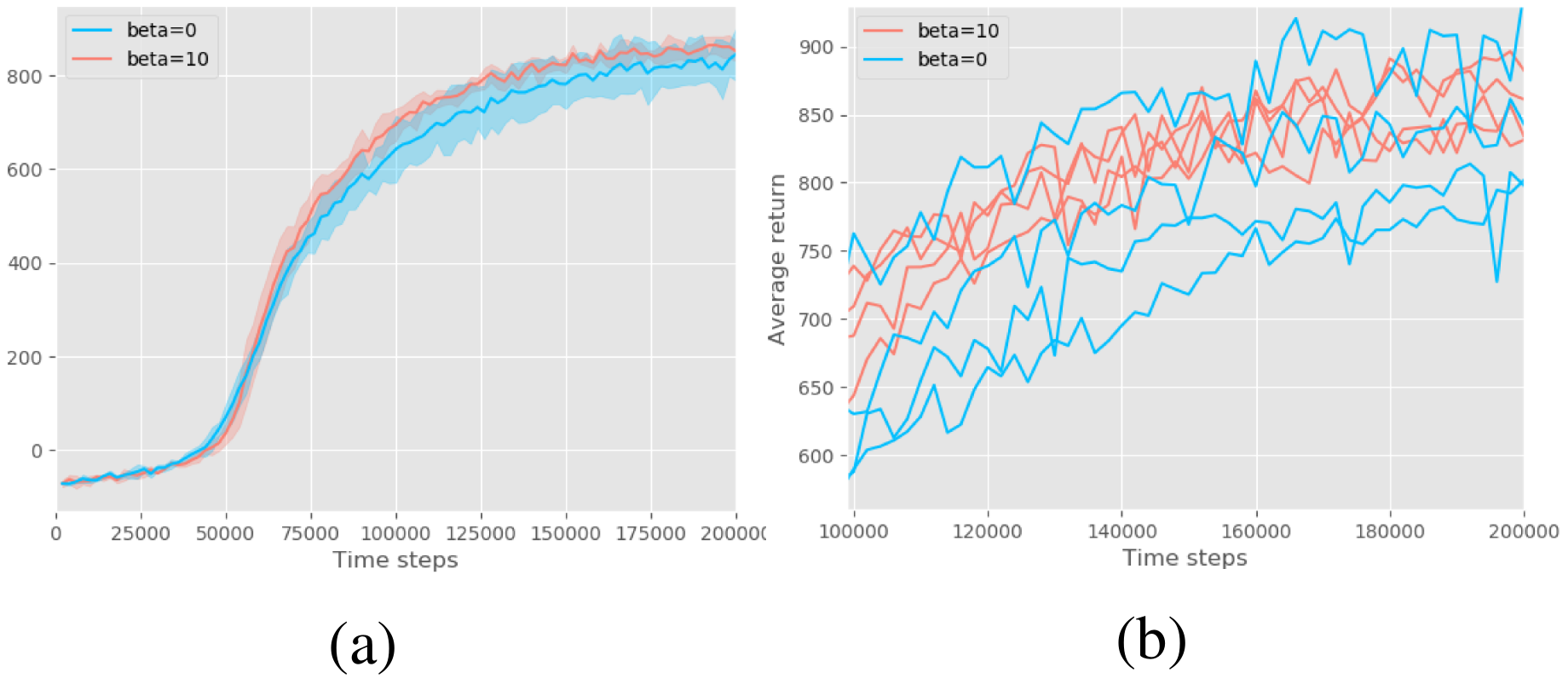}
		\caption{(a): The development of average return with $\beta=10$ and $\beta=0$. (b): The performance after 1e5 time-step training with different random seeds.}
		\label{fig:exploration}
	\end{figure}

	The results are shown in Figure \ref{fig:abalation} and \ref{fig:exploration}. Setting $\alpha$ as $0.5$ or $0.75$ achieves the best final performance and the best robustness with 200K samples. Note that a large $\alpha$ may result in poorer performance in the early stage, because the uncertainty is high in the early stage and a large $\alpha$ tends to choose a small step size when uncertainty is high. Using $\beta=10$ can improve the performance (larger mean and smaller standard deviation), which demonstrate the effectiveness of uncertainty-aware exploration.

	\subsection{Robustness Analyses}
	We demonstrate the excellent robustness of POMBU in two
	ways. First, we evaluate algorithms in noisy environments. In these environments, we add Gaussian noise to the observation with the standard deviation $\sigma$. This noise will affect the accuracy of the learned models. Second, we evaluate algorithms in long-horizon tasks. In these tasks, models need to generate long trajectories, and the error is further exacerbated due to the difficulty of longterm predictions.

\begin{figure}[ht]
		\centering
		\includegraphics[width=.85\columnwidth]{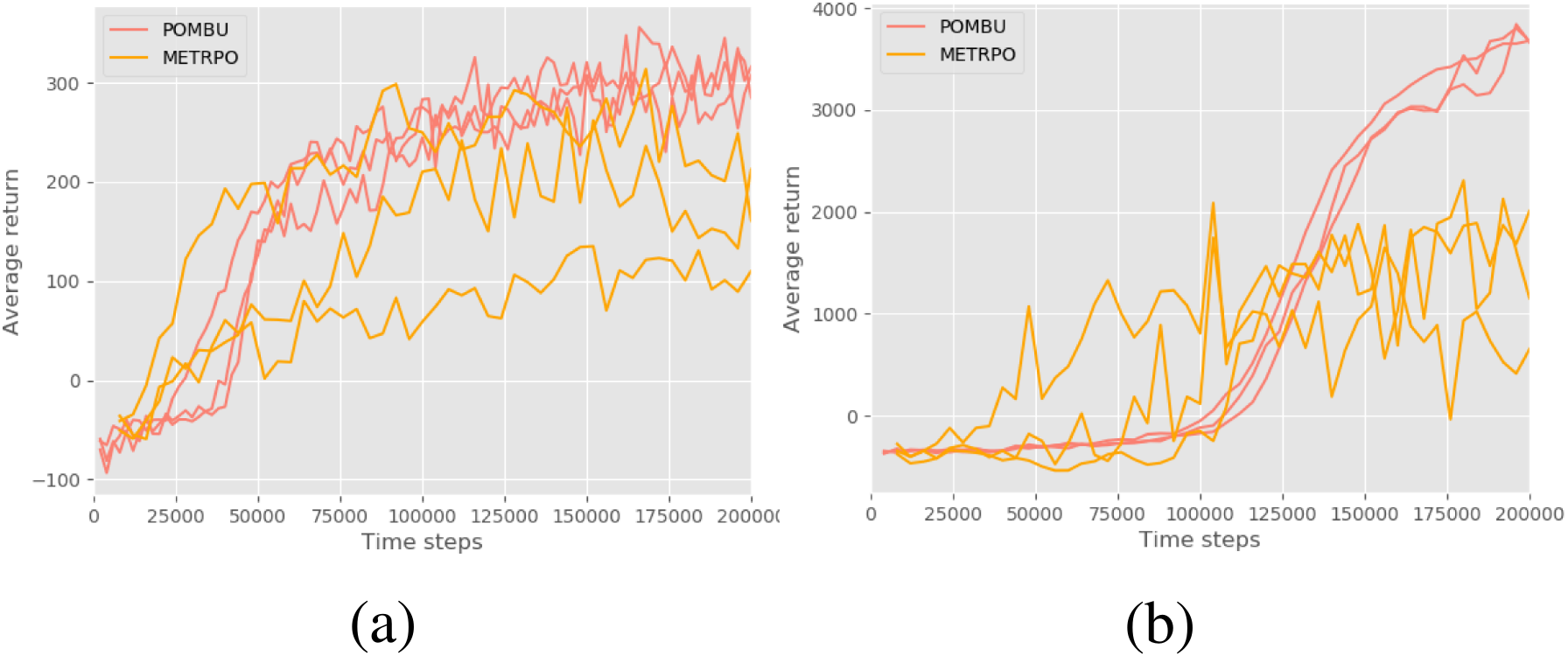}
		\caption{The training curves of POMBU and METRPO with different random seeds. (a) Comparison in a noisy Cheetah task ($\sigma=0.1$). (b) Comparison in a long-horizon Cheetah task ($H=1000$). }
		\label{fig:robust}
	\end{figure}

	We report the results in Figure \ref{fig:robust}. Experiments show that our algorithm achieves similar performance with different random seeds, while the performance of METRPO varies greatly with the random seeds. Moreover, in Figure \ref{fig:robust}, the worst performance of POMBU beats the best of METRPO. This implies that our method has promising robustness, even in noisy environments and long-horizon environments.

	\section{Conclusion}
	In this work, we propose a Policy Optimization method with Model-Based Uncertainty (POMBU), which is a novel uncertainty-aware model-based algorithm. This method estimates uncertainty using a model ensemble and then optimizes policy Conservatively considering the uncertainty. Experiments demonstrate that POMBU can achieve comparable asymptotic performance with SAC and PPO while using much fewer samples. Compared with other model-based methods, POMBU is robust and can achieve better performance. We believe that our approach will bring new insights into model-based reinforcement learning.
	An enticing direction for further work is the combination of our uncertainty estimation method with other kinds of models like Bayesian neural networks. Another exciting direction is to modify other advanced model-based algorithms like STEVE and PETS  using our uncertainty estimation method. 
	
	\fontsize{9.5pt}{10.5pt}
	\selectfont

	\setcounter{lemma}{0}
	\setcounter{theorem}{0}
	\setcounter{section}{0} 
	
	\newpage
	
	{
		{
			\centering\LARGE\bf Deep Model-Based Reinforcement Learning
			via Estimated Uncertainty and Conservative Policy Optimization
			
			Appendix
			\par
		}
	}
	\renewcommand\thesection{\Alph{section}}
	\vspace{20pt}
	In this appendix, we first present the detailed proof of all the lemmas and theorems. Then, we provide an algorithm to estimate the uncertainty in Q-values in long-horizon tasks. Moreover, we provide a empirical comparison between our method and UBE. Finally, we provide the details of our experiments settings and additional experimental results.
	
	\section{Proof}
	In this suction, We provide all the proof mentioned in the body of our paper.
	
	\subsection{Proof of Lemma 1}
	\begin{lemma}
		Under Assumption 1 and 2, for any $s \in \mathcal{S}$ and $a \in \mathcal{A}$, we have
		\begin{align*}
		&\mathbb{D}_\mathcal{M}\left[Q^h_{sa}\right] \leq u^{h}_{sa} + \sum_{s^\prime, a^\prime}\pi_{s^\prime a^\prime}\bar{P}_{sas^\prime}\mathbb{D}_\mathcal{M}\left[Q^{h+1}_{s^\prime a^\prime}\right],\\
		&\mbox{where } u^h_{sa} = \mathbb{D}_{\mathcal{M}}\left[R_{sa} + \sum_{s^\prime}P_{sas^\prime}\bar{V}^{h+1}_{s^\prime}\right].
		\end{align*}
	\end{lemma}
	\emph{Proof. } Let $\sum_i a_i = 1$ and each $X_i$ is a random variable. By using Jensen's inequality , we have
	\begin{align}
	\mathbb{D}\left[\sum_{i} a_i X_i\right] &= \mathbb{E}\left[\left(\sum_{i} a_i X_i - \mathbb{E}\left[\sum_{i} a_i X_i\right]\right)^2\right] \notag\\
	&= \mathbb{E}\left[\left(\sum_{i} a_i \left(X_i - \mathbb{E}\left[X_i\right]\right)\right)^2\right] \notag\\
	&\leq  \mathbb{E}\left[\sum_{i} a_i \left(X_i - \mathbb{E}\left[X_i\right]\right)^2\right] \notag\\
	&= \sum_{i} a_i \mathbb{D}\left[X_i\right]. \label{INEQ1}
	\end{align}
	By applying the inequation (\ref{INEQ1}) to the Bellman equation, we have 
	\begin{align}
	\mathbb{D}_{\mathcal{M}}\left[V^h_s\right] =\mathbb{D}_{\mathcal{M}}\left[\sum_{a}\pi_{sa}Q^h_{sa}\right] \leq \sum_{a}\pi_{sa}\mathbb{D}_{\mathcal{M}}\left[Q^h_{sa}\right].
	\label{INEQ2}
	\end{align}
	By using the law of total variance, we have 
	\begin{align}
	\mathbb{D}_{\mathcal{M}}\left[Q^h_{sa}\right]\notag 
	&=\mathbb{D}_{\mathcal{M}}\left[R_{sa} + \sum_{s^\prime}P_{sas^\prime} V^{h+1}_{s^\prime}\right]\notag\\
	&=\mathbb{D}_{P_{sa},R_{sa}}\left[\mathbb{E}\left[\left.R_{sa} + \sum_{s^\prime}P_{sas^\prime} V^{h+1}_{s^\prime}\right|P_{sa}, R_{sa}\right]\right] + \mathbb{E}_{P_{sa},R_{sa}}\left[\mathbb{D}\left[\left.R_{sa} + \sum_{s^\prime}P_{sas^\prime} V^{h+1}_{s^\prime}\right|P_{sa}, R_{sa}\right]\right] \label{EQ1}
	\end{align}
	Because Assumption 1 and 2 implies that 
	$
	\mathbb{E}\left[V^{h+1}_{s^\prime} \left.|P_{sa}, R_{sa}\right.\right] = \mathbb{E}_\mathcal{M}\left[V^{h+1}_{s^\prime}\right]
	$ when $s^\prime \neq s$
	, we have
	\begin{align}
	&\quad \  \mathbb{D}_{P_{sa},R_{sa}}\left[\mathbb{E}\left[\left.R_{sa} + \sum_{s^\prime}P_{sas^\prime} V^{h+1}_{s^\prime}\right|P_{sa}, R_{sa}\right]\right] \notag\\
	&=\mathbb{D}_{P_{sa},R_{sa}}\left[R_{sa} + \sum_{s^\prime}P_{sas^\prime}\mathbb{E}\left[\left. V^{h+1}_{s^\prime}\right|P_{sa}, R_{sa}\right]\right]\notag\\
	&=\mathbb{D}_{P_{sa},R_{sa}}\left[R_{sa} + \sum_{s^\prime}P_{sas^\prime}\bar{V}_{s^\prime}^{h+1}\right] = u^l_{sa}\label{INEQ3}
	\end{align}
	By using the inequation (\ref{INEQ1}), we have
	\begin{align}
	&\quad \ \mathbb{E}_{P_{sa},R_{sa}}\left[\mathbb{D}\left[\left.R_{sa} + \sum_{s^\prime}P_{sas^\prime} V^{h+1}_{s^\prime}\right|P_{sa}, R_{sa}\right]\right] \notag\\
	&\leq \mathbb{E}_{P_{sa},R_{sa}}\left[\sum_{s^\prime}P_{sas^\prime}\mathbb{D}\left[\left. V^{h+1}_{s^\prime}\right|P_{sa}, R_{sa}\right]\right]\notag\\
	&=\sum_{s^\prime}\bar{P}_{sas^\prime}\mathbb{D}\left[\left. V^{h+1}_{s^\prime}\right|P_{sa}, R_{sa}\right]\notag\\
	&=\sum_{s^\prime}\bar{P}_{sas^\prime}\mathbb{D}_\mathcal{M}\left[V^{h+1}_{s^\prime}\right],\label{INEQ4}
	\end{align}
	where the last step holds because $V^{h+1}_{s^\prime}$ is independent of $P_{sa}, R_{sa}$ when $s^\prime \neq s$ according to Assumption 1 and 2. Combining \ref{INEQ2}, \ref{EQ1}, \ref{INEQ3} and \ref{INEQ4}, we obtain the Lemma 1.
	\hfill $\blacksquare $

	\subsection{Proof of Theorem 1}
	\begin{theorem}
		Under Assumption 1 and 2, for any policy $\pi$, there exists a unique solution $U$ satisfying the following equation:
		\begin{align}
		&U^h_{sa} = u^h_{sa} + \sum_{s^\prime, a^\prime}\pi_{s^\prime a^\prime}\bar{P}_{sas^\prime}U^{h+1}_{s^\prime a^\prime}
		\end{align}
		for any $(s,a)$ and $h=1,2,\dots,H$, where $U^{H+1}=\textbf{0}$, and furthermore $U \geq \mathbb{D}_\mathcal{M}\left[Q\right]$ pointwise.
	\end{theorem}
	\emph{Proof. }First, the solution of $U^H$ exists and is unique because $U^{H+1} = \textbf{0}$ and $U^H$ is a linear combinations of $U^{H+1}$. Moreover, we know that 
	$
	\mathbb{D}_\mathcal{M}\left[Q^H_{sa}\right] \leq u^H_{sa} = U^H_{sa}.
	$
	
	Then, there exists a unique solution of $U^{H-(i+1)}$ if there exists a unique solution of $U^{H-i}$ because $U^{H-(i+1)}$ is a linear combinations of $U^{H-i}$ and $u^{H-(i+1)}$. Additionally, by using Lemma 1, we have
	$$
	\mathbb{D}_\mathcal{M}\left[Q^{H-(i+1)}_{sa}\right] \leq u^{H-(i+1)}_{sa} + \sum_{s^\prime, a^\prime}\pi_{s^\prime a^\prime}\bar{P}_{sas^\prime}\mathbb{D}_\mathcal{M}\left[Q^{H-i}_{s^\prime a^\prime}\right] \leq u^{H-(i+1)}_{sa} + \sum_{s^\prime, a^\prime}\pi_{s^\prime a^\prime}\bar{P}_{sas^\prime}U^{H-i}_{s^\prime a^\prime} = U^{H-(i+1)}_{sa}
	$$
	if $U^{H-i} \geq \mathbb{D}_\mathcal{M}\left[Q^{H-i}\right]$ pointwise. 
	
	Finally, we obtain Theorem 1 by induction.
	\hfill $\blacksquare $
	
	\subsection{Proof of Theorem 2}
	\begin{theorem}
		For arbitrary $\left(U\right)_1$, if
		$$
		\left(U^h_{sa}\right)_{i+1} = \left(u^h_{sa}\right)_i + \sum_{s^\prime, a^\prime}\pi_{s^\prime a^\prime}\bar{P}_{sas^\prime}\left(U^{h+1}_{s^\prime a^\prime}\right)_i,
		$$
		for any $(s,a)$, $h=1,2,\dots,H$ and $i \geq 1$, where $\left(U^{H+1}\right)_i=\textbf{0}$ and $\left(u\right)_i$ converges to $u$ pointwise, we have $\left(U\right)_i$ converges to $U$ pointwise.
	\end{theorem}
	\noindent\emph{Proof. }$\left(U^H\right)_i$ is converges to $U^H$ because $\left(u^H\right)_i$ converges to $u^H$ and $\left(U^H\right)_i = (u^H)_i$. 
	
	For any $j>0$, if $\left(U^{H-j+1}\right)_i$ converges to $U^{H-j+1}$, $\left(U^{H-j}\right)_i$ converges to $U^{H-j}$ with the assumption $\left(u^{H-j}\right)_i$ converges to $u^{H-j}$ because $\left(U^{H-j}\right)_i$ is a linear combinations of $\left(U^{H-j+1}\right)_i$ and $\left(u^{H-j}\right)_i$. 
	
	Then, we obtain the conclusion by induction. \hfill $\blacksquare $
	
	\subsection{Proof of Theorem 3}
	
	\begin{theorem}
		Under the assumption 1, 2 and 3, $U^h_{sa}$ is a tighter upper bound of $\mathbb{D}_\mathcal{M}\left[Q^h_{sa}\right]$ than $B^h_{sa}$.
	\end{theorem}
	\noindent\emph{Proof. } Here, we only show that
	$
	U\leq B
	$
	pointwise (see UBE \cite{o2017uncertainty} for the proof that $B$ is an upper bound of uncertainty).
	
	Because $\sum_{s^\prime}\mathbb{E}_\mathcal{M}\bar{P}_{sas^\prime}=1$, by using inequation (\ref{INEQ2}), we have
	\begin{align}
	\mathbb{D}_{\mathcal{M}}\left[\sum_{s^\prime}P_{sas^\prime} \bar{V}^{h+1}_{s^\prime}\right] 
	&=\mathbb{D}_{\mathcal{M}}\left[\sum_{s^\prime}\frac{\bar{P}_{sas^\prime}}{\bar{P}_{sas^\prime}}P_{sas^\prime} \bar{V}^{h+1}_{s^\prime}\right]\notag\\
	&\leq\sum_{s^\prime}\bar{P}_{sas^\prime}\mathbb{D}_{\mathcal{M}}\left[\frac{P_{sas^\prime}}{\bar{P}_{sas^\prime}} \bar{V}^{h+1}_{s^\prime}\right]\notag\\
	&=\sum_{s^\prime}\mathbb{D}_{\mathcal{M}}\left[\bar{P}_{sas^\prime} \bar{V}^{h+1}_{s^\prime}\right]/\bar{P}_{sas^\prime}\notag\\
	&\leq\sum_{s^\prime}\mathbb{D}_{\mathcal{M}}\left[\bar{P}_{sas^\prime} Q_{\mbox{max}}\right]/\bar{P}_{sas^\prime}\notag\\
	&=\left(Q_{\mbox{max}}\right)^2 \sum_{s^\prime}\mathbb{D}_{\mathcal{M}}\left[P_{sas^\prime}\right]/\bar{P}_{sas^\prime}.
	\label{tight_INEQ}
	\end{align}
	
	Under the Assumption 1, 2 and 3, we have 
	\begin{align*}
	u^h_{sa} &= \mathbb{D}_{\mathcal{M}}\left[R_{sa} + \sum_{s^\prime}P_{sas^\prime}\bar{V}^{h+1}_{s^\prime}\right] \\
	&= \mathbb{D}_{\mathcal{M}}\left[R_{sa}\right] + \mathbb{D}_{\mathcal{M}}\left[\sum_{s^\prime}P_{sas^\prime}\bar{V}^{h+1}_{s^\prime}\right]\\ &\leq
	\mathbb{D}_{\mathcal{M}}\left[R_{sa}\right] +
	\left(Q_{\mbox{max}}\right)^2 \sum_{s^\prime}\mathbb{D}_{\mathcal{M}}\left[P_{sas^\prime}\right]/\bar{P}_{sas^\prime} \\
	&= \nu^h_{sa}.
	\end{align*}
	
	Then $U^{H} = u^h \leq \nu^h \leq B^{H}$ pointwise. If $U^{H-i} \leq B^{H-i}$ pointwise, we have
	$$
	U^{H-(i+1)}_{sa} = u^{H-(i+1)}_{sa} + \sum_{s^\prime, a^\prime}\pi_{s^\prime a^\prime}\bar{P}_{sas^\prime}U^{H-i}_{s^\prime a^\prime} \leq \nu^{H-(i+1)}_{sa} + \sum_{s^\prime, a^\prime}\pi_{s^\prime a^\prime}\bar{P}_{sas^\prime}B^{H-i}_{s^\prime a^\prime} = B^{H-(i+1)}_{sa}.
	$$ 
	Moreover we have $U^{H+1}=B^{H+1}=\mathbf{0}$. 
	
	Then, we obtain the conclusion by induction.
	\hfill $\blacksquare$
	
	\subsection{Proof of Theorem 4}
	\begin{theorem}
		\begin{align*}
		\left.L_{sr}(\theta)\right|_{\theta=\theta_{\text{old}}} &= \left.J(\pi_{\theta}) - J(\pi_{\theta_{\text{old}}})\right|_{\theta=\theta_{\text{old}}},\\
		\left.\nabla_\theta L_{sr}(\theta)\right|_{\theta=\theta_{\text{old}}} &= \nabla_\theta\left.\left(J(\pi_{\theta}) - J(\pi_{\theta_{\text{old}}})\right)\right|_{\theta=\theta_{\text{old}}}.
		\end{align*}
	\end{theorem}
	
	\noindent\emph{Proof. }Note that 
	$$A^h_{{\text{old}}}(s,a) = \mathbb{E}_{\tau\sim (\mathcal{M}, \pi_{\theta})}\left[\left.R_{s^ha^h}+V^{h+1}_{{\text{old}}}(s^{h+1})-V^h_{{\text{old}}}(s^h)\right|s^h=s,a^h=a\right].
	$$ 
	Therefore, we have
	\begin{align}
	\mathbb{E}_{\tau\sim (\mathcal{M},\pi_{\theta})}\left[\sum_{h=1}^{H}A^h_{{\text{old}}}(s^h,a^h)\right] 
	&=\mathbb{E}_{\tau\sim (\mathcal{M}, \pi_{\theta})}\left[\sum_{h=1}^{H}\left(R_{s^ha^h}+V^{h+1}_{{\text{old}}}(s^{h+1})-V^h_{{\text{old}}}(s^h)\right)\right] \notag\\
	&=\mathbb{E}_{\tau\sim (\mathcal{M}, \pi_{\theta})} \left[-V^1_{{\text{old}}}(s^1) + \sum_{h=1}^{H}R_{s^ha^h}\right] \notag\\
	&=-\mathbb{E}_{\tau\sim (\mathcal{M}, \pi_{\theta})}\left[V^1_{{\text{old}}}(s^1)\right] + \mathbb{E}_{\tau\sim (\mathcal{M}, \pi_{\theta})}\left[ \sum_{h=1}^{H}R_{s^ha^h}\right]\notag\\
	&=-J(\pi_{\text{old}})+J(\pi_\theta).
	\label{lemmaeq1}
	\end{align}
	When $\pi_{\text{old}} = \pi_\theta$, we have 
	\begin{align}
	L_{sr}(\theta) &=  \mathbb{E}_{\tau\sim (\mathcal{M}, \pi_{\theta_{\text{old}}})}\left[\sum_{h=1}^{H}r_\theta(s^h,a^h)A^h_{{\text{old}}}(s^h,a^h)\right]
	=\mathbb{E}_{\tau\sim (\mathcal{M}, \pi_{\theta})}\left[\sum_{h=1}^{H}A^h_{{\text{old}}}(s^h,a^h)\right]\notag
	= J(\pi_\theta) - J(\pi_{\text{old}}).
	\end{align}
	Additionally, we have
	\begin{align}
	\nabla_\theta\mathbb{E}_{\tau\sim (\mathcal{M}, \pi_{\theta})}\left[\sum_{h=1}^{H}A^h_{{\text{old}}}(s^h,a^h)\right]
	&=\nabla_\theta\sum_{h=1}^{H}\mathbb{E}_{\tau\sim (\mathcal{M}, \pi_{\theta})}\left[A^h_{{\text{old}}}(s^h,a^h)\right] \notag\\
	&=\nabla_\theta\sum_{h=1}^{H}\mathbb{E}_{\tau\sim (\mathcal{M}, \pi_{\theta_{\text{old}}})}\left[\left(\prod_{l=1}^hr_\theta(s^l,a^l)\right)\times A^h_{{\text{old}}}(s^h,a^h)\right]\notag\\
	&=\sum_{h=1}^{H}\mathbb{E}_{\tau\sim (\mathcal{M}, \pi_{\theta_{\text{old}}})}\left[\left(\nabla_\theta\prod_{l=1}^{h-1}r_\theta(s^l,a^l)\right)\times r_\theta(s^h,a^h)\times A^h_{{\text{old}}}(s^h,a^h)\right] + \notag\\
	&\qquad \sum_{h=1}^{H}\mathbb{E}_{\tau\sim (\mathcal{M}, \pi_{\theta_{\text{old}}})}\left[\left(\prod_{l=1}^{h-1}r_\theta(s^l,a^l)\right)\times \nabla_\theta r_\theta(s^h,a^h)\times A^h_{{\text{old}}}(s^h,a^h)\right].
	\label{lemmaeq2}
	\end{align} 
	When $\pi_{\text{old}} = \pi_\theta$, we have
	\begin{align}
	&\mathbb{E}_{\tau\sim (\mathcal{M}, \pi_{\theta_{\text{old}}})}\left[\left(\nabla_\theta\prod_{l=1}^{h-1}r_\theta(s^l,a^l)\right)\times r_\theta(s^h,a^h)\times A^h_{{\text{old}}}(s^h,a^h)\right]\notag\\
	=&\mathbb{E}_{\tau\sim (\mathcal{M}, \pi_{\theta_{\text{old}}})}\left[\left(\nabla_\theta\prod_{l=1}^{h-1}r_\theta(s^l,a^l)\right)\times\left(Q^h_{{\text{old}}}(s^h,a^h)-V^h_{{\text{old}}}(s^h)\right) \right]\notag \\
	=&\mathbb{E}_{\tau\sim (\mathcal{M}, \pi_{\theta_{\text{old}}})}\left[\left(\nabla_\theta\prod_{l=1}^{h-1}r_\theta(s^l,a^l)\right)\times\left(V^h_{{\text{old}}}(s^h)-V^h_{{\text{old}}}(s^h)\right) \right]\notag \\
	=&0
	\label{lemmaeq3}
	\end{align}
	and
	\begin{align}
	&\sum_{h=1}^{H}\mathbb{E}_{\tau\sim (\mathcal{M}, \pi_{\theta_{\text{old}}})}\left[\left(\prod_{l=1}^{h-1}r_\theta(s^l,a^l)\right)\times \nabla_\theta r_\theta(s^h,a^h)\times A^h_{{\text{old}}}(s^h,a^h)\right] \notag\\
	=&\sum_{h=1}^{H}\mathbb{E}_{\tau\sim (\mathcal{M}, \pi_{\theta_{\text{old}}})}\left[\nabla_\theta r_\theta(s^h,a^h)\times A^h_{{\text{old}}}(s^h,a^h)\right] \notag\\
	=&\nabla_\theta\mathbb{E}_{\tau\sim (\mathcal{M}, \pi_{\theta_{\text{old}}})}\left[\sum_{h=1}^{H} r_\theta(s^h,a^h) A^h_{{\text{old}}}(s^h,a^h)\right] \notag\\
	=&\nabla_\theta L_{surr}(\theta).
	\label{lemmaeq4}
	\end{align}
	Combine equations (\ref{lemmaeq1}), (\ref{lemmaeq2}), (\ref{lemmaeq3}) and (\ref{lemmaeq4}), we have $
	\left.\nabla_\theta L_{sr}(\theta)\right|_{\theta=\theta_{\text{old}}} = \nabla_\theta\left.\left(J(\pi_{\theta}) - J(\pi_{\theta_{\text{old}}})\right)\right|_{\theta=\theta_{\text{old}}}$. 
	\hfill $\blacksquare $
	
	\newpage
	
	\section{Uncertainty Estimation with a Discount Factor}
	When considering a discount factor $\gamma \in (0,1]$, we train $\tilde{V}_\phi$ by minimizing $\ell_2$ loss function  
	\begin{align}
	L_v(\phi) =\mathbb{E}_{\tau\sim(\hat{\mathcal{M}}, \pi)}\left[\frac{1}{H}\sum_{h=1}^{H}\left\|\tilde{V}_\phi(s^h)- \sum_{l=h}^{H}\gamma^{l-h}\hat{R}_{s^la^l}\right\|_2^2\right].
	\end{align} and estimate uncertainty using the algorithm \ref{gamma}.
	
	\begin{algorithm}[ht]
		\caption{Uncertainty Estimation for Q-values with a discount factor $\gamma$}
		\label{gamma}
		\SetAlgoNoLine 
		\SetKwInOut{Input}{\textbf{Input}}\SetKwInOut{Output}{\textbf{Output}}
		\Input{A approximate value function $\tilde{V}_\phi$; An ensemble model $\{f_{w_1}, f_{w_2}, \dots, f_{w_K} \}$; A discount factor $\gamma$; A trajectory $\tau=(s_1, a_1, \dots, s_H, a_H)$\;\\}
		\Output{Estimates of $(\sqrt{U^1_{s_1a_1}}, \dots, \sqrt{U^H_{s_Ha_H}})$;\\}
		\BlankLine
		$D^{H+1}=0$\;
		\For{$i = H,H-1,\dots,1$}{
			\For{$j = 1,2,\dots,K$}{
				$(s_j, r_j) = f_{w_j}(s^i,a^i)$\;
				$q_j = r_j + \gamma \tilde{V}_\phi(s_j)$\;
			}
			$q = \frac{1}{K}\sum_{j=1}^{K}q_j$\;
			$d^i = \frac{1}{K}\sum_{j=1}^{K}(q_j-q)^2$\;
			$D^{i} = d^i + \gamma^2D^{i+1}$\;
		}
		\Return $(\sqrt{D^1}, \sqrt{D^2}, \dots, \sqrt{D^H})$;
	\end{algorithm}

	\section{Empirical Comparison with Different Horizons}
\begin{figure}[ht]
		\centering
		\includegraphics[width=200pt]{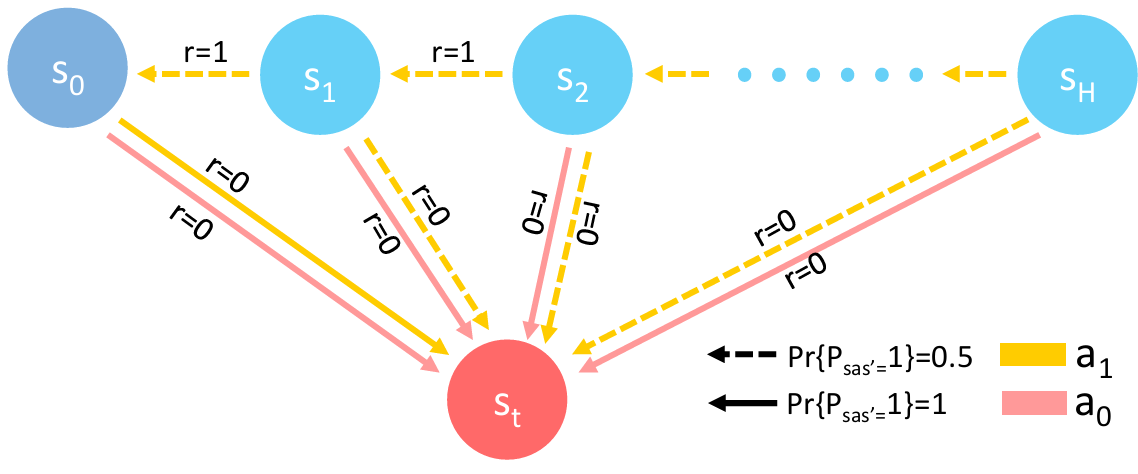}
		\caption{An intuitive definition of the probability space.}
		\label{p_space}
	\end{figure}

	Here, we compare our uncertainty estimation method with UBE empirically. We consider a simple posterior of models denoted by  $(\varOmega,\mathcal{F},Pr)$. Each MDP $\mathcal{M} \in \varOmega$ has a same finite action space $\{a_0, a_1\}$ and a same finite state space $\{s_t, s_0, s_1, \dots, s_H\}$, where $s_t$ is the terminal state.
	We use a deterministic ensemble $\{f_1, f_2\}$ to represent the probability measure function like discussed in Section 3. The ensemble are defined by $f_1(s, a) = (s_t, 0)$ and
	$$
	f_2(s, a) =
	\begin{cases}
	(s_{i-1}, 1), &\mbox{if }a = a_1 \mbox{ and } s=s_i \mbox{ for any } i>0,\\
	(s_t, 0), &\mbox{otherwise}.
	\end{cases}
	$$
	More intuitively, we represent those possible MDPs as the picture \ref{p_space}.
	
\begin{figure}[ht]
		\centering 
		\begin{subfigure}{0.23\textwidth}
			\includegraphics[width=0.95\textwidth]{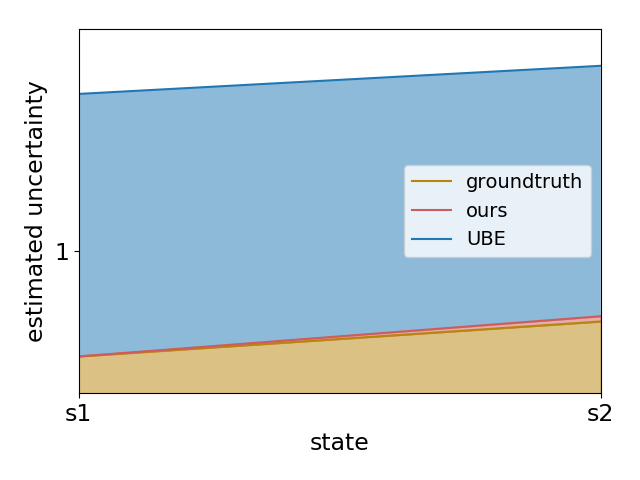}
			\caption{H=2}
		\end{subfigure} 
		\begin{subfigure}{0.23\textwidth}
			\includegraphics[width=0.95\textwidth]{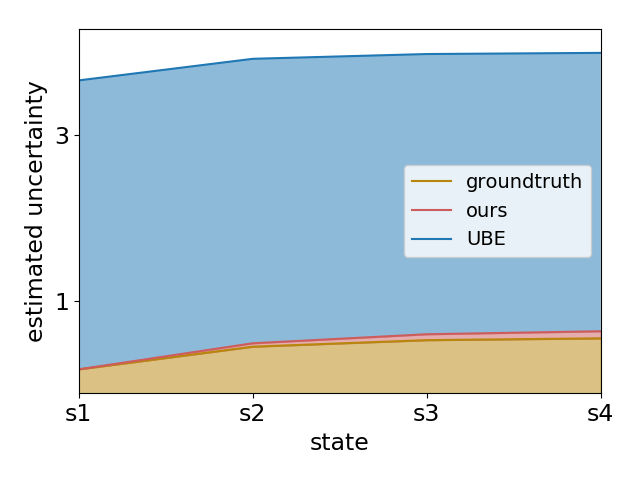}
			\caption{H=4}
		\end{subfigure}
		\begin{subfigure}{0.23\textwidth}
			\includegraphics[width=0.95\textwidth]{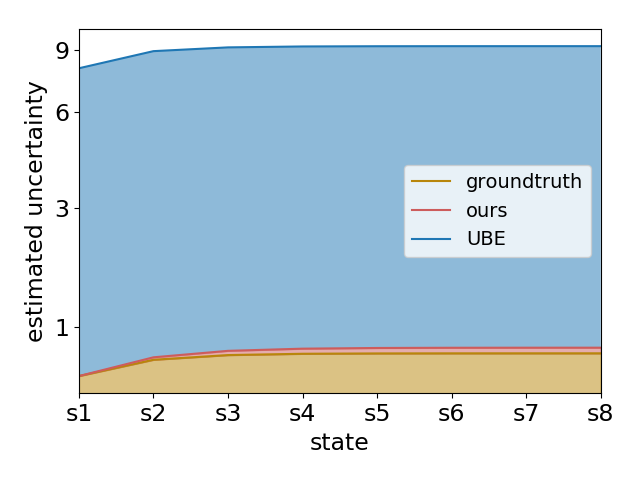}
			\caption{H=8}
		\end{subfigure}
		\begin{subfigure}{0.23\textwidth}
			\includegraphics[width=0.95\textwidth]{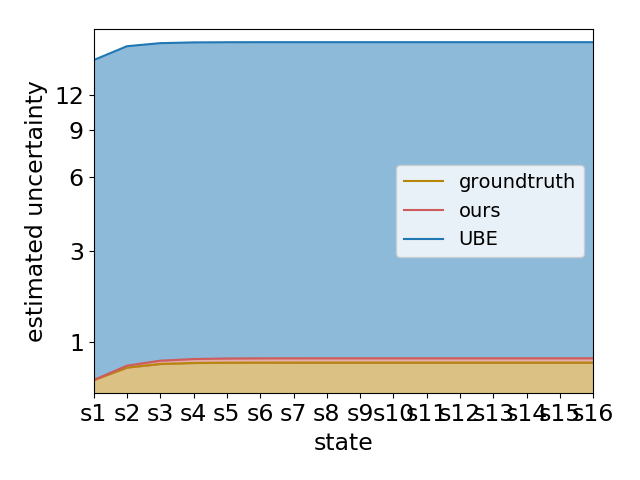}
			\caption{H=16}
		\end{subfigure}
		\caption{The comparison between estimated uncertainties with different horizon. }
		\label{fig:uncertainty}
	\end{figure}

	We compute the value and the uncertainty under a policy $\pi$. This policy satisfies that $\pi_{sa}=0.5$ for any state $s$, action $a$.
	We compare $U$ (estimated by our method) and $B$ (estimated by UBE) with the groundtruth $\mathbb{D}_\mathcal{M}\left[Q\right]$. The results are shown in Figure \ref{fig:uncertainty}. It shows that both our method and UBE esimate upper bounds of the true uncertainty. The upper bound estimated by our method is much tighter, especially when using a long horizon.
	\newpage
	
	\section{Experiments}
	We describe the details of our experiments and provide additional experimental results in this part.
	
	\subsection{Effectiveness in 2D-point 3D-point Environments}
	We first introduce the 2D-point environment. Our 2D-point environment is similar to that defined by Ignasi Clavera \cite{clavera2018model}.
	\begin{itemize}
		\item \textbf{State space: }$\mathbb{R}^2$. 
		\item \textbf{Action space: }$[-0.1, 0.1]^2$. 
		\item \textbf{Initial state distribution: }Uniform distribution over $[-2, 2]^2$.
		\item \textbf{Transition function: }The distribution of the next state is $\delta(s+a)$, where $\delta$ means a dirac distribution.
		\item \textbf{Reward function: }$R_{sa}=-\|s+a\|^2_2$.
		\item \textbf{Horizon: }$H=30$.
	\end{itemize}
	The 3D-point environment is defined similarly to 2D-point environment while the dimensionality of its state space and action space is 3.
	
	Then we introduce how we estimate the error and value. We sample 5000 states from the initial state distribution and select actions using a deterministic policy $\pi:\mathcal{S}\rightarrow A$ (parameterize by a random initialized neural network). We use an ensemble containing 5 models, each of which has the hidden layer sizes (64,64). The approximate value function $\tilde{V}_w$ has the layer sizes (32,32). We collect 3000 real-world trajectories to train the models using mini-batch size 500 and learning rate 0.0002. We collect 100 virtual trajectories to train the value function per epoch using mini-batch size 200 and learning rate 0.00005.
	
\begin{figure}[h]
		\centering 
		\begin{subfigure}{0.25\textwidth}
			\includegraphics[width=0.95\textwidth]{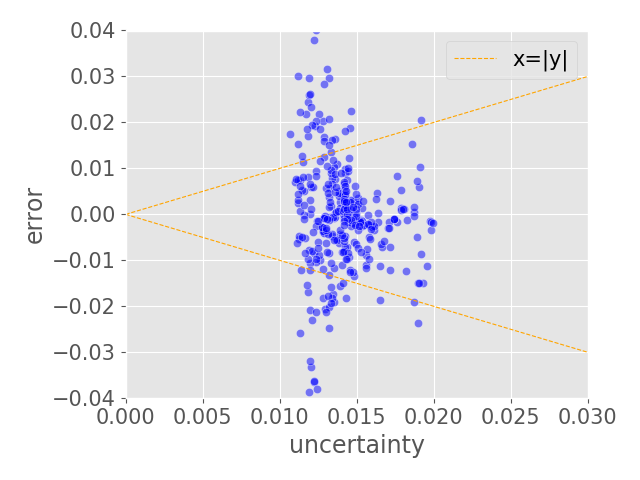}
			\caption{after 0 epochs}
		\end{subfigure} 
		\begin{subfigure}{0.245\textwidth}
			\includegraphics[width=0.95\textwidth]{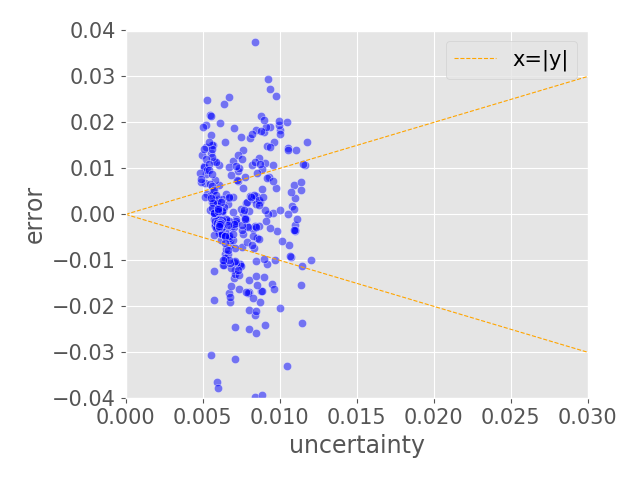}
			\caption{after 5 epochs}
		\end{subfigure}
		\begin{subfigure}{0.245\textwidth}
			\includegraphics[width=0.95\textwidth]{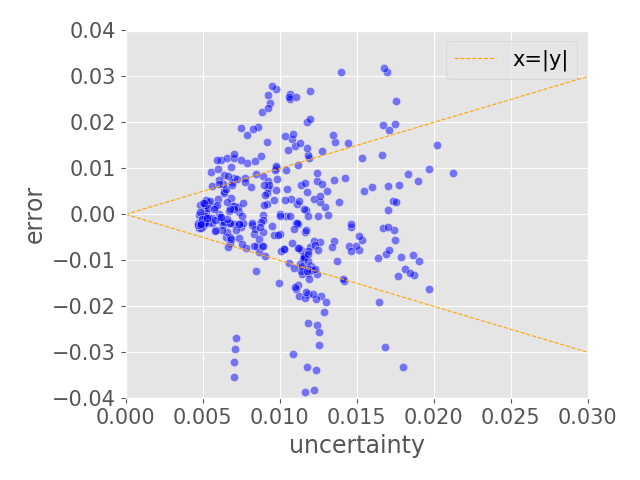}
			\caption{after 10 epochs}
		\end{subfigure}
		\begin{subfigure}{0.245\textwidth}
			\includegraphics[width=0.95\textwidth]{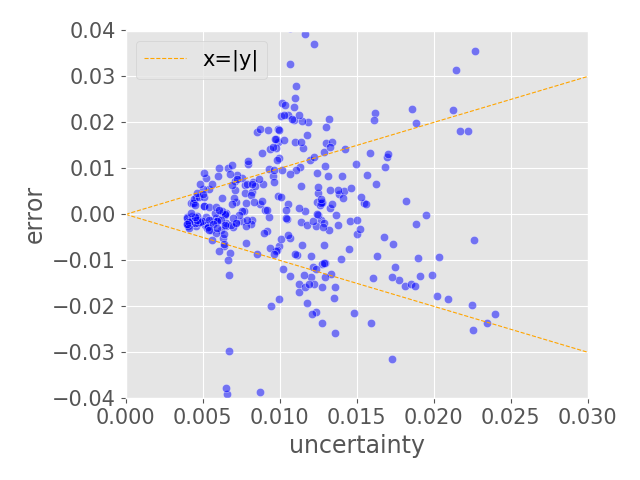}
			\caption{after 15 epochs}
		\end{subfigure}
		\caption{Relation between error and uncertainty. We show the results after different number of epochs (training $\tilde{V}_w$).}
		\label{fig:scatter}
	\end{figure}

	We also show the relation between error and uncertainty using scatters (sample 500 pairs of data) in Figure \ref{fig:scatter} (2-D point).
	
	\subsection{Mujoco Environments}
	We use the environments adopted from Tingwu Wang's implements \cite{1907.02057}. The details are as follow.
	
	\begin{table}[h]
		\centering
		\begin{tabular}{c  c  c  c}
			\whline
			\textbf{Environment Name} & \textbf{Observation Space Dimension } & \textbf{Action Space Dimension} & \textbf{Horizon}\\ \hline
			Ant & 28 & 8 & 200 \\ 
			Walker2D & 17 & 6 & 200  \\ 
			HalfCheetah & 17 & 6 & 200  \\ 
			Swimmer & 8 & 2 & 200  \\ \whline
		\end{tabular}
		
		\caption{Dimensions of observation and action space, and horizon length.}
		
	\end{table}
	\begin{table}[h]
		\centering
		\begin{tabular}{c  c}
			\whline
			\textbf{Environment Name} & \textbf{Reward Function}\\ \hline
			Ant & $\dot{x}-0.1\|a\|^2_2 - 3.0 \times (z-0.57)^2$ \\ 
			Walker2D & $\dot{x}-0.1\|a\|^2_2 - 3.0 \times (z-1.3)^2$  \\ 
			HalfCheetah & $\dot{x}-0.1\|a\|^2_2$ \\ 
			Swimmer & $\dot{x}-0.0001\|a\|^2_2$ \\ \whline
		\end{tabular}
		
		\caption{Reward functions. In this table, $\dot{x}$ denotes the $x$ direction velocity, $a$ is the action and $z$ denotes the height.}
		
	\end{table}

	\subsection{Implement and Hyperparameter Setting}
	First we provide the hyperparameter setting of our method for the experiments in Section 6.2.
	
	\textbf{Dynamics Model: }
	In all models, we used weight normalization and ReLu nonlinearities. We composed an ensemble using 5 models. We used the Adam optimizer with a batch-size of 1000 and learning
	rate 0.0002 to minimize the $L_m(\phi)$ of each model. For each model, we randomly split the dataset using a 4-to-1 ratio for training and validation datasets. Except for the first iteration, all models were trained with warm starts using the parameters of the previous iteration. We trained each model individually and stopped the training individually until the validation loss has not decreased by 25 epochs (we validated the training every 5 epochs). We also normalized the input and output to train the models.
	
	\textbf{Policy Optimization: }We represented the policy $\pi_{\theta}$ using a 2-hidden-layer neural network with hidden sizes (64,64) and tanh nonlinearities for all the environments. We used the $\epsilon=0.15$ to clip the objective function and used $\alpha=0.5$ for all short horizon tasks. For training exploration policy, we used $\beta=10$. All policies were optimized by Adam optimizer with a linear decreasing learning rate. For each step of update, we trained the policy in 10 epochs.
	
	\textbf{Value and Uncertainty Estimation: }We represented the value function $\tilde{V}_w$ using a 2-hidden-layer neural network with hidden sizes (64,64) and tanh nonlinearities for all the environments. We normalized the rewards using the standard deviation of returns. We trained the approximate value function using Adam optimizer with a linear decreasing learning rate. For each step, we trained the approximate value function in 10 epochs.
	
	In the long horizon tasks, we show the different hyperparameters in the Tabel \ref{hyperpara_r}. For long-horizon HalfCheetah task, we set $\alpha=0.75$, and for long-horizon Ant task, we set $\alpha=0.25$. The other hyperparameters are same as the original.
	
	\begin{table}[h]
		\centering
		\begin{tabular}{c  c  c  c  c}
			\whline
			\textbf{Hyperparameter } & \textbf{Ant } & \textbf{Walker2D} & \textbf{HalfCheetah} & \textbf{Swimmer}\\ \hline
			\textbf{hidden layers (model)}& (512,512,512) & (512,512) & (512,512) & (512,512) \\ 
			\textbf{number of real-world trajectories per iteration} & 20 & 20 & 10 & 10  \\ 
			\textbf{number of updates per iteration} & 30 & 30 & 20 & 20  \\ 
			\textbf{number of iterations} & 150 & 100 & 100 & 50 \\
			\textbf{number of virtual trajectories per update} & 500 & 500 & 200 & 200  \\
			{\bfseries mini-batch size ($\pi_\theta$ and $\bar{V}_w$)} & 1250 & 1250 & 500 & 500  \\
			{\bfseries learning rate in $i$th iter. ($\pi_\theta$ and $\bar{V}_w$)} & $\frac{160-i}{1.6\times10^6}$ & $\frac{110-i}{1.1\times10^6}$ & $\frac{110-i}{1.1\times10^6}$ & $\frac{60-i}{0.6\times10^6}$  \\
			\whline
		\end{tabular}
		
		\caption{Hyperparameter setting in short horizon tasks (H=200).}
		\label{hyperpara}
		
	\end{table}

	\begin{table}[h]
		\centering
		\begin{tabular}{c  c  c }
			\whline
			\textbf{Hyperparameter } & long-horizon tasks ($H=1000$) \\ \hline
			\textbf{number of real-world trajectories per iteration} & 4   \\ 
			\textbf{number of updates per iteration} & 30   \\ 
			\textbf{number of iterations} & 50 \\
			\textbf{number of virtual trajectories per update} & 50 \\
			{\bfseries mini-batch size ($\pi_\theta$ and $\bar{V}_w$)} & 500   \\
			{\bfseries learning rate in $i$th iter. ($\pi_\theta$ and $\bar{V}_w$)}  & $\frac{60-i}{0.6\times10^6}$\\
			\whline
		\end{tabular}
		
		\caption{Different hyperparameter setting in long horizon tasks (H = 1000).}
		\label{hyperpara_r}
	\end{table}

	Then we introduce our baselines. We chose the state-of-the-art policy optimization algorithms (including both model-base and model-free methods that optimize parameterized policies) as our baselines. We did not compare our method to model-based meta policy optimization (MBMPO) \cite{clavera2018model} in short horizon tasks because of its high computation cost. It is worth noting that the probabilistic ensembles with trajectory sampling (PETS) \cite{chua2018deep} select actions via model predictive control (MPC) instead of a parameterized policy. Therefore, PETS is another kind of algorithm different from our method. We also did not compare our method to PETS in short horizon tasks. There are the implements that we used to draw figure 2.
	\begin{itemize}
		\item \textbf{Soft actor critic (SAC): }\cite{haarnoja2018soft} A state-of-the-art model-free algorithm for continuous control in terms of sample efficiency. We use the implement from rlkit\footnote{https://github.com/vitchyr/rlkit}.
		\item \textbf{Proximal policy optimization (PPO): }\cite{schulman2017proximal} A simple and efficient model-free algorithm. We use Tingwu Wang's implement \cite{1907.02057}.
		\item \textbf{Stochastic Lower Bounds Optimization (SLBO): }\cite{luo2018algorithmic} A deep model-based algorithm with a theoretical guarantee of monotone improvement to a local maximum of the expected reward. We use the office implement from the author.
		\item \textbf{Model-Ensemble Trust Region Policy Optimization (METRPO): }\cite{kurutach2018model} A deep model-based algorithm that optimizes policy via trust region policy optimization algorithm using imaginary data. We use Tingwu Wang's implement \cite{1907.02057}.
		\item \textbf{Model-Ensemble Proximal Policy Optimization (MEPPO): }A standard deep model-based algorithm that optimizes policy via proximal policy optimization algorithm using imaginary data. we implement it based on the code of our method by setting $\alpha=0$ and $\beta=0$. 
	\end{itemize}
	
	For all those model-based baselines, we used the default of other hyperparameters. For our method and MEPPO, we evaluate the performance using the average return of 20 trajectories.

	\subsection{How does POMBU Work?}
\begin{figure}[ht]
		\centering
		\vspace{-15pt}
		\includegraphics[width=0.85\textwidth]{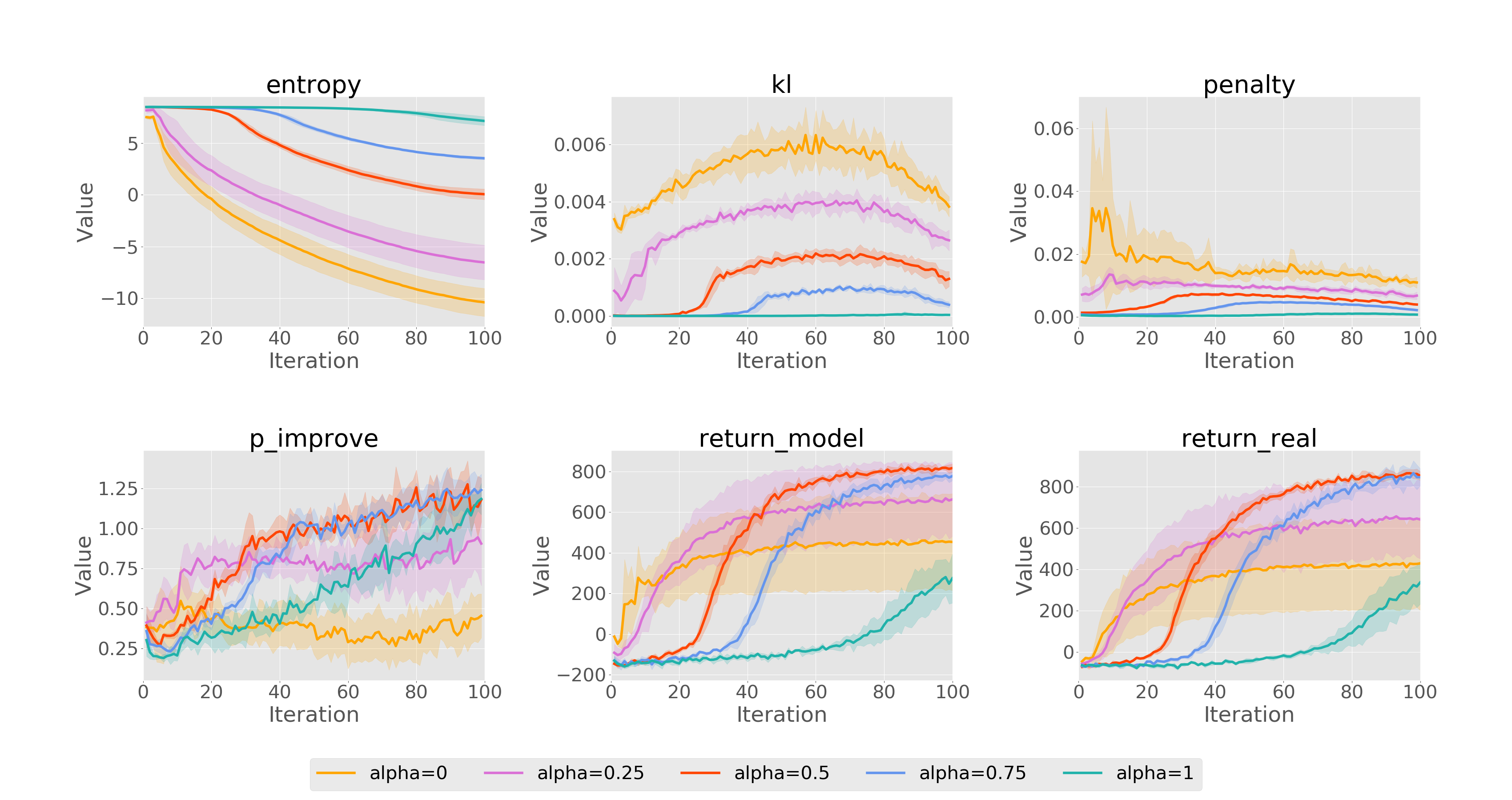}
		\caption{The details during training using different $\alpha$ in HalfCheetah task ($H=200$).}
		\label{fig:trainning}
	\end{figure}

	To further show how our algorithm work, we provide the details during training in Figure \ref{fig:trainning}. In this picture, we show the development of different kinds of values with the increase of iterations. 
	\begin{itemize}
		\item[] \textbf{entropy}: The entropy of policy. Here, we report the value at the end of each iteration.
		\item[] \textbf{kl}: The averaged Kullback-Leibler (KL) divergence between the post-update policy and the pre-update policy. During a iteration, we update the policy multiple times and estimate the averaged KL divergence for each update. A large value of KL means a large the step size of update. We report the average for each iteration.
		\item[] \textbf{penalty}: $\mathbb{E}_{\tau\sim (\mathcal{M}, \pi_{\theta_{\text{old}}})}\left[\sum_{h=1}^{H}\left|r_\theta(s^h,a^h)-1\right|\sqrt{U^h_{sa}}\right]$. The estimated uncertainty over the surrogate objective $L_{sr}(\theta)$. For each iteration, we report the average during this iteration.
		\item[] \textbf{p\_improve}: $L_{clip}(\theta)$. An estimated probability of performance improvement. For each iteration, we report the average during this iteration.
		\item[] \textbf{return\_model}: The average returns of virtual trajectories generated by the models. We report the value at the end of each iteration.
		\item[] \textbf{return\_real}: The average returns of real-world trajectories. We report the value at the end of each iteration.
	\end{itemize}
	With the increase of $\alpha$, the step size decreases. In the early stage, the value of kl is approximately 0 due to the high inaccuracy of models. This show the conservatism of our method. The small step size leads to a slow rate of descent in entropy. This ensures better exploration after multiple iterations. Small $\alpha$ results in higher uncertainty over the surrogate objective and lower probability of performance improvement. When using $\alpha=0$, the return\_model is unstable and much higher than return\_real in the early stage, which means overfitting. Using $\alpha > 0$ can significantly alleviate this overfitting, even using $\alpha=0.25$. It is worth noting that, updating policy with small stepsize---for example, kl$\leq0.001$ when $\alpha=0.75$, which is \textbf{10 times} smaller than most implements of TRPO choose---can achieve promising sample efficiency.
	
	\section{Robust Analysis}
	Here, we provide more results for robust analysis.

	\begin{figure}[!h]
		\centering
		\begin{subfigure}{0.33\textwidth}
			\includegraphics[width=0.95\textwidth]{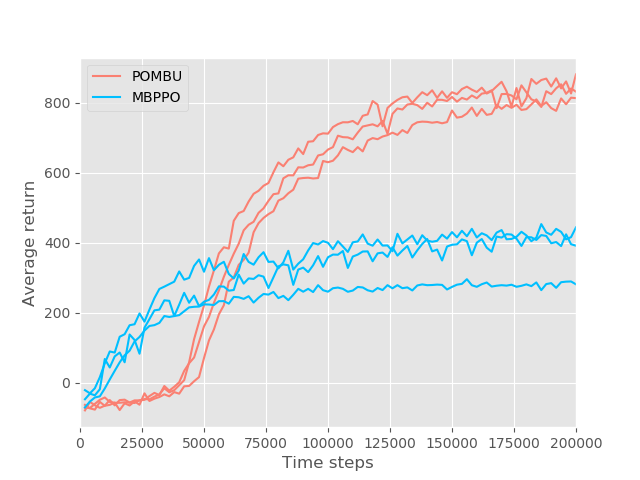}
			\caption{We compare POMBU to MEPPO in a noisy environment ($\sigma=0.01$).}
		\end{subfigure}
		\begin{subfigure}{0.33\textwidth}
			\includegraphics[width=0.95\textwidth]{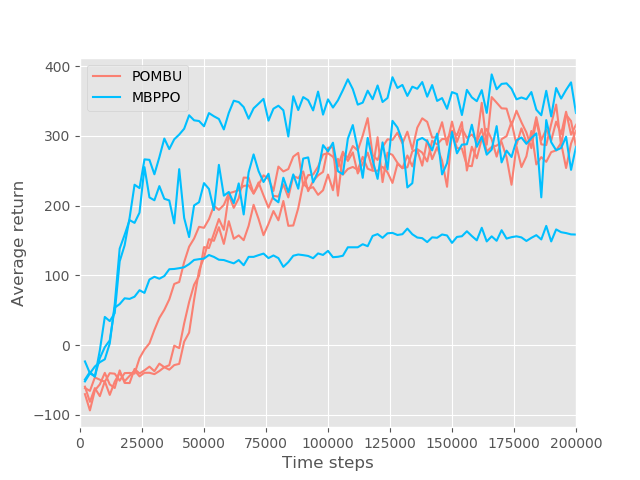}
			\caption{We compare POMBU to MEPPO in a noisy environment ($\sigma=0.1$).}
		\end{subfigure} 
		\begin{subfigure}{0.33\textwidth}
			\includegraphics[width=0.95\textwidth]{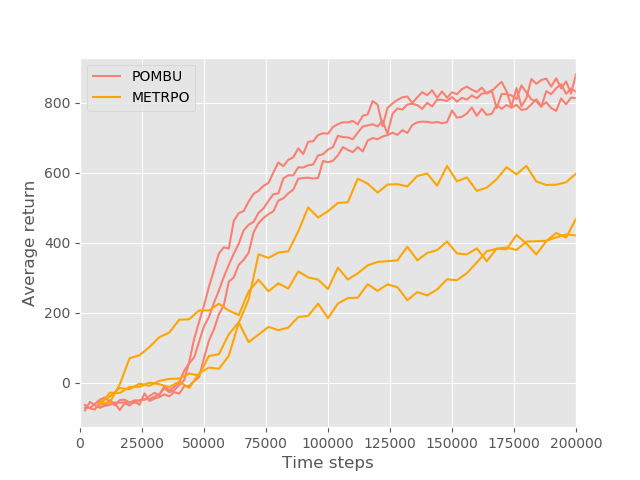}
			\caption{We compare POMBU to METRPO in a noisy environment ($\sigma=0.01$).}
		\end{subfigure} 
		\caption{Robustness comparisons in noisy HalfCheetah tasks .}
		\label{noise_rest}
	\end{figure}
	
	\begin{figure}[!h]
		\centering
		\begin{subfigure}{0.48\textwidth}
			\includegraphics[width=0.95\textwidth]{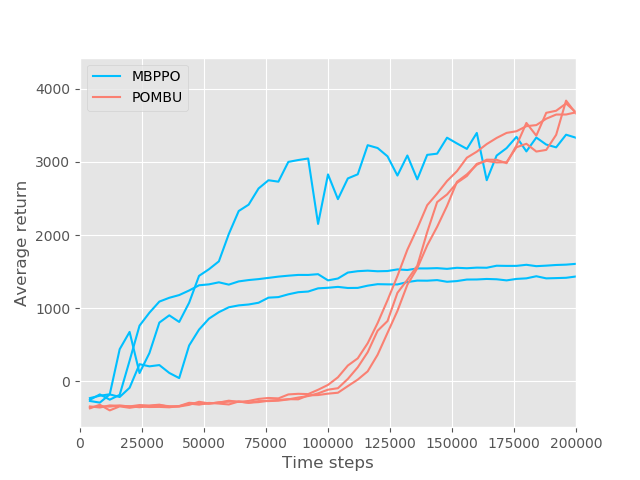}
			\caption{Long-horizon HalfCheetah environment ($H=1000$).}
		\end{subfigure}
		\begin{subfigure}{0.48\textwidth}
			\includegraphics[width=0.95\textwidth]{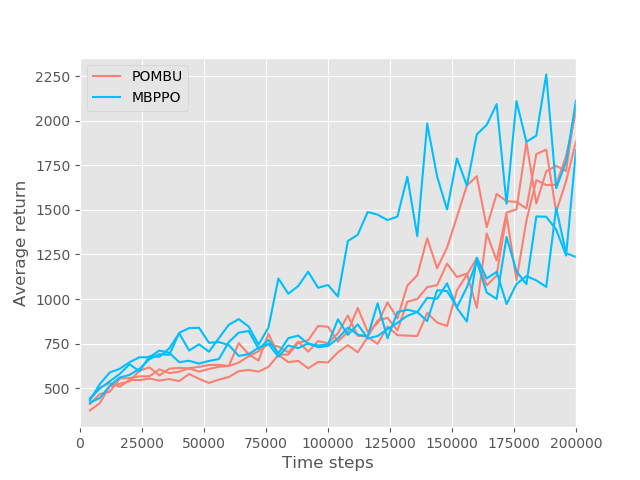}
			\caption{Long-horizon Ant environment ($H=1000$).}
		\end{subfigure} 
		
		\begin{subfigure}{0.48\textwidth}
			\includegraphics[width=0.95\textwidth]{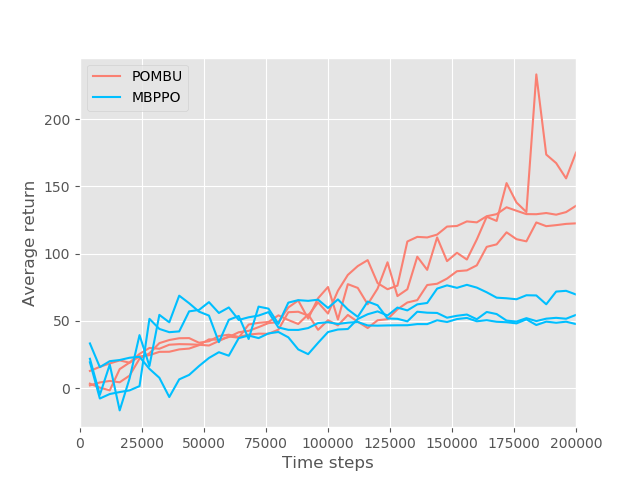}
			\caption{Long-horizon Swimmer environment ($H=1000$).}
		\end{subfigure}
		\begin{subfigure}{0.48\textwidth}
			\includegraphics[width=0.95\textwidth]{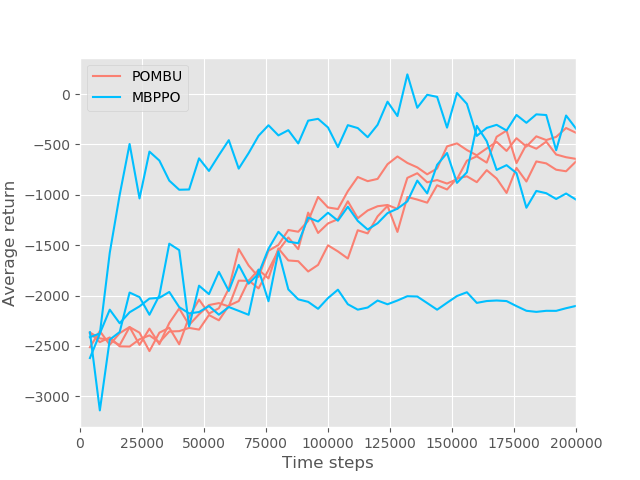}
			\caption{Long-horizon Walker2D environment ($H=1000$).}
		\end{subfigure} 
		\caption{Robustness comparisons between our method and MEPPO in long-horizon tasks.}
		\label{long_tasks}
	\end{figure}

\end{document}